\documentclass[letterpaper]{article} % DO NOT CHANGE THIS
\usepackage{aaai2026}  % DO NOT CHANGE THIS
\usepackage{times}  % DO NOT CHANGE THIS
\usepackage{helvet}  % DO NOT CHANGE THIS
\usepackage{courier}  % DO NOT CHANGE THIS
\usepackage[hyphens]{url}  % DO NOT CHANGE THIS
\usepackage{graphicx} % DO NOT CHANGE THIS
\usepackage{natbib}  % DO NOT CHANGE THIS AND DO NOT ADD ANY OPTIONS TO IT
\usepackage{caption} % DO NOT CHANGE THIS AND DO NOT ADD ANY OPTIONS TO IT
\usepackage{algorithm}
\usepackage{algorithmic}

\usepackage{newfloat}
\usepackage{listings}
\DeclareCaptionStyle{ruled}{labelfont=normalfont,labelsep=colon,strut=off} % DO NOT CHANGE THIS
\floatstyle{ruled}
\newfloat{listing}{tb}{lst}{}
\floatname{listing}{Listing}
\usepackage{tcolorbox}
\usepackage{enumitem}
\usepackage{ragged2e}

\usepackage{booktabs}
\usepackage{multirow}
\usepackage{xcolor} % ATTENTION MAYBE SHOULD REMOVE
\usepackage{subcaption}
\usepackage{amsmath,amsfonts,bm,amsthm,amssymb}
\newcommand{\dependencyarrow}{\ensuremath{\rightarrow}}

\title{Interactive Evaluation of Large Language Models for Multi-Requirement Software Engineering Tasks}
\author {
    Dimitrios Rontogiannis,\textsuperscript{\rm 1}
    Maxime Peyrard,\textsuperscript{\rm 2}
    Nicolas Baldwin \textsuperscript{\rm 3},
    Martin Josifoski \textsuperscript{\rm 4, *},
    Robert West \textsuperscript{\rm 3},
    Dimitrios Gunopulos \textsuperscript{\rm 1}
}

\affiliations {
    \textsuperscript{\rm 1} Department of Informatics and Telecommunications, National and Kapodistrian University of Athens, Athens, Greece \\
    \textsuperscript{\rm 2} Université Grenoble Alpes, CNRS, Grenoble INP, LIG \\
    \textsuperscript{\rm 3} EPFL \\
    \textsuperscript{\rm 4} FAIR at Meta \\
    dronto@di.uoa.gr, maxime.peyrard@univ-grenoble-alpes.fr,  nicolas.baldwin@epfl.ch, martinjosifoski@meta.com, robert.west@epfl.ch, dg@di.uoa.gr\\
    \textsuperscript{*} work done while at EPFL
}

\begin{document}

    \maketitle

\begin{abstract}

Standard single-turn, static benchmarks fall short in evaluating the nuanced capabilities of Large Language Models (LLMs) on complex tasks such as software engineering. In this work, we propose a novel interactive evaluation framework that assesses LLMs on multi-requirement programming tasks through structured, feedback-driven dialogue.
Each task is modeled as a requirement dependency graph, and an ``interviewer'' LLM, aware of the ground-truth solution, provides minimal, targeted hints to an ``interviewee'' model to help correct errors and fulfill target constraints. This dynamic protocol enables fine-grained diagnostic insights into model behavior, uncovering strengths and systematic weaknesses that static benchmarks fail to measure.
We build on DevAI, a benchmark of 55 curated programming tasks, by adding ground-truth solutions and evaluating the relevance and utility of interviewer hints through expert annotation.
Our results highlight the importance of dynamic evaluation in advancing the development of collaborative code-generating agents.
\end{abstract}

%     \link{Code + Dataset}{http://bit.ly/4fiMSkU}

\section{Introduction}

The integration of Large Language Models (LLMs) into software development has transformed coding from a solitary, linear process into a dynamic, iterative collaboration. Modern tools like ChatGPT \cite{schulman2022chatgpt}, DeepSeek \cite{deepseekai2025deepseekv3technicalreport}, and AI-first IDEs such as Cursor exemplify this shift, where developers no longer simply request code but refine it through multiturn dialogues. Feedback, whether clarifications, corrections, or incremental constraints, has become the scaffold for progress, allowing models to adapt to ambiguities, edge cases, and evolving requirements. Yet, despite this reality, the prevailing benchmarks continue to evaluate LLMs as static single-turn code generators, ignoring the very interactions that define their practical utility.

\paragraph{The Gap in Current Evaluation}

Current evaluation paradigms for software engineering problems suffer from two critical misalignments with real-world software workflows. (i) First, they treat tasks as monolithic problems \cite{chen2021evaluatinglargelanguagemodels, hendrycks2021measuringcodingchallengecompetence}, ignoring their compositional nature. For example, building a recommendation system requires strict dependencies: data loading $\rightarrow$ feature engineering $\rightarrow$ model training $\rightarrow$ API exposure. Yet static evaluations force models to ``guess correctly'' on the first attempt, conflating understanding of requirements with luck in initial output. This penalizes models for early errors (e.g., data loading) and obscures their ability to recover in downstream steps (e.g., model training), even though real development often involves debugging partial solutions. (ii) Second, while recent work explores interactive evaluation \cite{wang2023mint,pan2025benchmarks}, these efforts rely on shallow feedback (e.g., binary correctness checks) or unstructured hints, failing to capture the directed repair behavior of human-AI collaboration. In practice, a model's value depends on its ability to adapt, say fixing a missing edge case after a developer's nudge, but benchmarks rarely measure this. The gap is systemic: without evaluating how LLMs leverage feedback to navigate dependencies or rectify cascading errors, we risk overestimating failures (where models could recover) or underestimating pitfalls (where models pass single-turn tests but reveal critical flaws when exposed to step-by-step refinement) - precisely the dynamics that define their practical utility.

\paragraph{Our framework: Interactive, Dependency-Grounded Assessment}

\begin{figure}[h]
    \centering
    \includegraphics[width=0.4\textwidth]{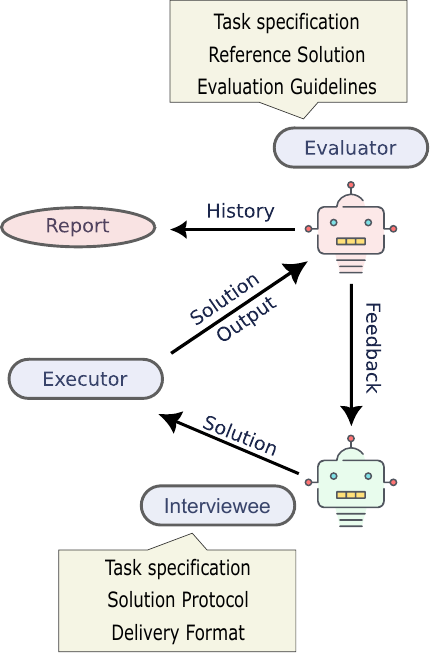}
    \caption{The Interactive Evaluation pipeline.}
    \label{fig:granular_pipeline}
\end{figure}

We propose a structured, feedback-driven evaluation framework (Figure \ref{fig:granular_pipeline}) for software engineering tasks. Each task is decomposed into a directed acyclic graph (DAG) of requirements, capturing the hierarchical dependencies between subtasks. A model is evaluated not only on its initial output but also on its ability to improve iteratively through targeted feedback loops. These hints are automatically generated by an LLM-based interviewer with access to ground truth solutions and task requirements. If a model fails an early subtask, we guide it past the error to assess its performance on subsequent steps - mirroring how human developers work around intermediate bugs to evaluate deeper functionality.

A critical design feature is our lightweight integration protocol. The framework exposes simple interfaces that allow any LLM to participate as either interviewer or interviewee with minimal adaptation. Researchers can evaluate new models by implementing just these basic interaction primitives, while still benefiting from the full power of our dependency-aware assessment pipeline. This modular design ensures wide applicability without compromising the richness of evaluation, enabling both controlled benchmarking and real-world deployment testing.

\paragraph{Contributions}

Our main contributions are:

\begin{enumerate}
    \item 
     A Dependency-Driven Interactive Evaluation Protocol: We introduce the first framework that jointly models software task decomposition and iterative feedback for LLM assessment. The framework's novel structure enables quantifying error propagation and recovery through guided feedback and easy integration via minimal interface requirements, allowing any LLM to participate as interviewer or interviewee with trivial adaptation.

    \item 
    An enhanced DevAI benchmark: We augment DevAI \cite{zhuge2024agentasajudgeevaluateagentsagents} with verified Ground-Truth solutions, using the original Agent-as-a-Judge methodology to ensure correctness. This extension enables guided, multi-stage evaluation through our framework's structured feedback mechanism, resulting in an improved benchmark that serves as both (i) an evaluation platform for our experiments and (ii) a reusable resource for future interactive assessment frameworks.

    \item 
    Our experiments reveal that failures in static evaluations become recoverable with targeted feedback, suggesting that single-turn benchmarks severely underestimate LLM capabilities. We also identify critical failure modes where models cannot effectively incorporate feedback, revealing limitations in their ability to refine solutions even with iterative guidance.
\end{enumerate}

By bridging the gap between static benchmarks and real-world software workflows, our work advances practical LLM evaluation for software engineering problems.

\section{Related Work} \label{related_work}

Traditional evaluations of LLMs rely on static benchmarks with fixed inputs and binary success criteria. While benchmarks such as HumanEval \cite{chen2021evaluatinglargelanguagemodels}, APPS \cite{hendrycks2021measuringcodingchallengecompetence}, and MBPP \cite{austin2021programsynthesislargelanguage} have driven rapid progress in code generation, they fail to capture the process-oriented, iterative nature of real-world problem solving. These benchmarks typically assess models based on functional correctness of output in a single shot setting, which assumes complete and unambiguous task specifications, an assumption that does not hold in most practical development scenarios. Extensions like CodeXGLUE \cite{lu2021codexgluemachinelearningbenchmark} and SWE-bench \cite{jimenez2024swebenchlanguagemodelsresolve} move towards more realistic evaluation tasks, such as bug fixing and issue resolution in real codebases. However, they still emphasize static correctness over dynamic reasoning, offering limited insight into how models handle ambiguity, adapt over time, or respond to developer intent.

To address these shortcomings, recent work has explored \textit{interactive evaluation}, where models are assessed over multiple turns with access to feedback or clarification. Human-in-the-loop setups such as iEval \cite{svikhnushina-etal-2022-ieval} and CheckMate \cite{collins2023evaluatinglanguagemodelsmathematics} demonstrate that interactivity reveals model competencies overlooked by static scoring, particularly in complex domains like mathematics or natural language understanding. These studies show that model performance can vary substantially when they are allowed to ask questions, request hints, or revise outputs based on critique. More scalable frameworks simulate interaction using LLMs both as agents and evaluators, as in IQA-Eval \cite{li2024iqaeval}, KIEval \cite{yu2024kievalknowledgegroundedinteractiveevaluation}, and medical roleplay systems \cite{liao2024automaticinteractiveevaluationlarge}, enabling broader experimentation without relying on human annotators. These frameworks highlight the potential of structured feedback to surface model behaviors that are otherwise invisible in single-turn evaluations.

Complementary to interactivity, \textit{adaptive evaluation} dynamically adjusts testing based on model responses. DyVal \cite{zhu2024dyvaldynamicevaluationlarge} and DyVal 2 extend this idea using reasoning graphs and skill-specific probes to isolate weaknesses and trace error propagation through multistep reasoning tasks. These tools allow for a more diagnostic view of model performance, showing not just whether a model fails, but how and why it fails across different cognitive skills. Similarly, AdaTest \cite{ribeiro-lundberg-2022-adaptive} and benchmark self-evolving frameworks \cite{wang2024benchmarkselfevolvingmultiagentframework} generate targeted adversarial examples to stress-test models under varying conditions. By continuously updating the test set in response to model behavior, these approaches create a moving target that reveals brittleness or blind spots that static benchmarks overlook.

In the software engineering domain, a few agent-based approaches have emerged to better reflect realistic development pipelines. Notably, Agent-as-a-Judge \cite{zhuge2024agentasajudgeevaluateagentsagents} evaluates LLMs on tasks involving interdependent components such as planning, execution, and evaluation. These methods begin to address compositionality and dependency tracking, yet often lack structured mechanisms for feedback-based refinement. They typically evaluate outputs at isolated checkpoints, without modeling how developer guidance might help correct or improve the model’s trajectory through a task.

Our work builds on this trajectory by introducing an interactive evaluation framework tailored to software engineering tasks. Unlike prior efforts that isolate interactivity, adaptivity, or software-specific evaluation, our approach unifies these aspects through requirement decomposition, dependency-aware scoring, and guided iterative feedback. This allows for a more granular and realistic assessment of how models reason, adapt, and improve in complex engineering workflows, capturing the collaborative dynamics that characterize human-AI co-programming.
\section{Problem Formulation} \label{sec:problem_formulation}

We define \textbf{Interactive Software Engineering Evaluation} as a multi-stage assessment framework designed to evaluate large language models through iterative refinement cycles guided by structured feedback. This approach specifically addresses complex, decomposable tasks characterized by three key properties: first, the presence of hierarchical dependencies among requirements; second, the potential need for incremental correction of partial solutions; and third, the necessity to evaluate both initial capability and adaptive improvement. The framework represents tasks as Directed Acyclic Graphs (DAGs) of requirements, where vertices correspond to verifiable subtasks, and edges encode functional dependencies.  

Unlike traditional binary evaluations, which assess success or failure on a task as a whole, interactive evaluation captures both the model's initial performance and its ability to refine and repair its output in response to minimal guidance. This approach aligns closely with real-world software engineering, where developers iteratively build and correct solutions in response to feedback.

\paragraph  {Structured Tasks with Hierarchical Dependencies} We focus on problems that consist of multiple interdependent requirements, where progress on earlier subtasks enables progress on later ones. Formally, let a task \( T \) be defined by a set of requirements \( R = \{ r_1, r_2, \dots, r_m \} \), with each \( r_j \) representing a subcomponent or constraint of the overall solution.

To capture the hierarchical and sequential structure of such tasks, we can model their dependencies as a DAG \( G = (R, E) \), where an edge \( (r_i, r_j) \in E \) indicates that requirement \( r_j \) can only be addressed after requirement \( r_i \) has been successfully completed. Let $P(r_j)$ denote the set of prerequisite requirements for $r_j$, and let $g \colon R \to \{0,1\}$ be the initial evaluation function that checks whether each requirement $r \in R$ is satisfied ($1$) or not ($0$). Then, \( r_j \) is evaluable only if all its parents are satisfied: $\forall r_i \in P(r_j), g(r_i) = 1$.

The effective (dependency-aware) evaluation score for the task is then defined as:

\begin{equation}
S_{G} = \frac{1}{m} \sum_{j=1}^{m} g(r_j) \cdot \mathbb{I} \left[ \forall r_i \in P(r_j),\ g(r_i) = 1 \right]
\label{eq:dag_initial_score}
\end{equation}

This formulation allows us to credit partial progress while respecting the task’s logical structure, avoiding overly coarse binary evaluations.

\paragraph{Guided Evaluation via Feedback}

Crucially, we are interested not just in how well a model performs on its first attempt, but in how effectively it improves when given feedback. In software engineering, a developer might suggest minimal edits (``rename this variable'', ``fix the off-by-one error''), guiding progress without solving the problem outright. We aim to replicate this process in evaluation.

Let \( R_{\text{fail}} \subseteq R \) be the set of requirements the model initially fails, and let \( H = \{ h_1, h_2, \dots, h_k \} \) be a minimal set of corrective hints provided by the evaluator. These hints serve as feedback for revision. The updated evaluation function \( g'_{H}(r_{j}) \) checks if the revised response meets $r_j$ given $H$.

We define the final interactive evaluation score as:

\begin{equation}
S_{G}' = \frac{1}{m} \sum_{j=1}^{m} g'_{H}(r_j) \cdot \mathbb{I} \left[ \forall r_i \in P(r_j),\ g'_{H}(r_i) = 1 \right]
\label{eq:dag_score_hints}
\end{equation}

By comparing \( S_{G} \) and \( S_{G}' \), we gain insight into a model’s capacity not just for initial accuracy, but for refinement—an essential skill in real-world applications. This interactive framework enables more efficient exploration of the solution space through feedback, aligning model evaluation with realistic software development workflows.

\begin{figure}[t]
    \centering
    \begin{tcolorbox}[
        colframe=black,
        colback=red!15,
        coltitle=black,
        title=\textbf{\textcolor{black}{Report Example}},
        fonttitle=\bfseries,
        boxrule=0.4mm,
        width=0.95\columnwidth,
        colbacktitle=blue!10,
        halign title=flush center,
        boxsep=2pt,
        left=2pt,
        right=2pt,
        top=2pt,
        bottom=2pt,
    ]
    \begin{enumerate}
        \item \textbf{Error Handling in Image Downloading}: The initial implementation of the \texttt{download\_image} function did not adequately handle connection errors...
        
        \item \textbf{URL Accessibility}: The model initially used a URL for the style image that resulted in a 404 error...
        
        \item \textbf{Logging and Feedback}: The model's initial logging for download attempts was minimal...
        
        \item \textbf{Code Organization}: While the code was well-structured...
        
        \item \textbf{Adaptability}: The model demonstrated good adaptability...
    \end{enumerate}
    
    Overall, the hints provided were instrumental...
    \end{tcolorbox}
    \caption{Example evaluation report}
    \label{fig:report_example}
\end{figure}

\paragraph{Post-Evaluation Report }

Following the interactive evaluation process, we generate a structured performance report to analyze the model’s strengths, weaknesses, and adaptability. Rather than providing a single aggregate score, this report captures multiple dimensions of the model’s behavior. It assesses problem-solving ability (e.g., whether the model can comprehend complex tasks and produce structured solutions), optimization awareness (e.g., consideration of time or space complexity), and, where applicable, code quality and organization. It also examines the model’s ability to recognize and correct its own mistakes, its responsiveness to minimal feedback, and its handling of ambiguity or incomplete information.

\section{Methodology}
\label{sec:methodology}

We introduce a structured methodology to evaluate LLMs on complex, structured software engineering tasks. The process consists of three stages: requirement extraction and initial evaluation, interactive refinement through feedback, and post-evaluation analysis.

\subsection{Ground Truth Construction, Requirement Extraction and Initial Evaluation}

Given a task \( T \), we construct a ground-truth solution \( S^* \) and define a set of core requirements \( R = \{ r_1, r_2, \dots, r_m \} \), representing essential aspects a correct solution must satisfy. These are structured as a DAG \( G = (R, E) \), where an edge \( (r_i, r_j) \in E \) indicates that \( r_j \) depends on the prior satisfaction of \( r_i \). We demonstrate our framework using the DevAI benchmark, which provides structured requirements but lacks Ground-Truth solutions.

To evaluate a model-generated solution \( S \), we segment it into chunks \( C = \{ c_1, c_2, \dots, c_n \} \) and embed both requirements and chunks using a sentence encoder \( f_{\text{enc}} \). For each requirement \( r_j \), we retrieve the most similar chunk \( c_k^* \) via cosine similarity. The pair \( (r_j, c_k^*) \) is then passed to an LLM-based classifier, which predicts whether the requirement is satisfied, conditioned on the satisfaction of its parent requirements in the DAG.

This initial evaluation procedure follows the \emph{Agent as a Judge} approach~\cite{zhuge2024agentasajudgeevaluateagentsagents}, and we adopt their judge implementation in our experiments.

\begin{figure}
    \centering
    \begin{tcolorbox}[
        colframe=black,
        colback=brown!30,
        coltitle=black,
        title=\textbf{\textcolor{black}{Interviewee: o3-mini, Problem: S26}},
        fonttitle=\bfseries,
        boxrule=0.4mm,
        width=\columnwidth,
        colbacktitle=blue!10,
        halign title=flush center,
        boxsep=2pt,
        left=2pt,
        right=2pt,
        top=2pt,
        bottom=2pt
    ]
    \justifying
    \emph{Your solution currently does not explicitly load the Electronics subset of the Amazon Reviews 2023 dataset using the \texttt{datasets} library as required. Instead, it reads from a local CSV or uses a dummy dataset fallback. To meet the requirement, please implement data loading in \texttt{src/data\_loader.py} using the \texttt{load\_dataset} function from the \texttt{datasets} library with the \texttt{"McAuley-Lab/Amazon-Reviews-2023"} dataset and \texttt{"raw\_review\_Electronics"} configuration, as shown in the reference solution. Also, please add explicit inline comments referencing the requirement for data loading and preprocessing steps.}
    \end{tcolorbox}
    \caption{Hint provided by the interviewer}
    \label{fig:hint_example}
\end{figure}

\subsection{Interactive Evaluation}

To measure a model's ability to improve its solution with guidance, we introduce an iterative evaluation loop. At each iteration \( t \), the model submits a revised solution \( S^{(t)} \), which is executed in a sandboxed Python environment to produce outputs \( O^{(t)} \) and errors \( E^{(t)} \) if any.

A separate LLM-based component, the \emph{interviewer} \( \mathcal{I} \), analyzes the current output, execution errors, the evaluation graph \( G \), and the ground-truth solution \( S^* \). Based on this, it generates a minimal set of natural language hints \( H^{(t)} \) intended to help the model correct its current deficiencies. These hints target specific failed requirements while preserving as much of the model’s original reasoning as possible. A concrete example of such hint can be seen in Figure \ref{fig:hint_example}, demonstrating how they guide iterative improvement without overcorrecting. Additional examples showing hint variation across different failure modes are provided in Appendix C.

The evaluated model receives \( H^{(t)} \) as input and produces an updated response \( S^{(t+1)} \). This loop continues until either all requirements are satisfied according to \( \mathcal{I} \), or a predefined maximum number of iterations is reached. 

At the end of the process, we compute the final interactive score using Equation~\eqref{eq:dag_score_hints}.

\subsection{Post-Evaluation Reporting}

Beyond correctness scores, we produce a qualitative report analyzing the model’s behavior throughout the evaluation trajectory \( \{ S^{(t)}, H^{(t)} \}_{t=1}^{T} \). This report is generated by an LLM-based analyzer \( \mathcal{R} \), which synthesizes insights about the model's reasoning process, adaptability, and robustness.

The analysis covers multiple dimensions, including problem-solving ability, sensitivity to feedback, optimization awareness (e.g., runtime or memory considerations), handling of ambiguity, and quality of code structure and organization. Rather than summarizing with a single metric, this report provides a structured breakdown of the model’s strengths and failure patterns, offering deeper insight into its underlying capabilities. The example in Figure \ref{fig:report_example} shows a typical report. Additional reports showcasing varied response patterns across different model architectures and task categories are available in Appendix B.

An overview of this multi-phase evaluation process is illustrated in Figure \ref{fig:granular_pipeline}, showing how model responses evolve through feedback and refinement.

\section{Experimental Setup: Benchmark and Models}
\begin{figure}[t!]
    \centering
    \includegraphics[width=\linewidth]{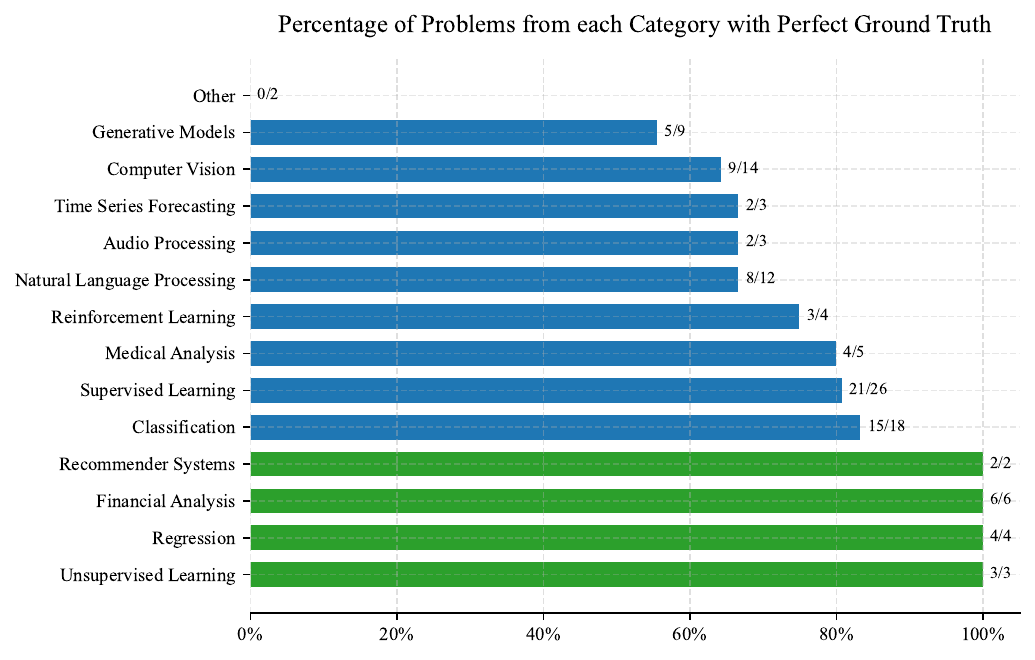}
    \caption{Percentage or problems with perfect ground truth accuracy by category.}
    \label{fig:perfect}
\end{figure}

For the Interactive Evaluation experiments, we utilize problems sourced from the \textbf{DevAI} benchmark, running all computations on an Apple M2 Pro system (12-core CPU with 8 performance/4 efficiency cores, 19-core GPU with Metal 3 acceleration, and 16GB unified memory). These software engineering problems span several machine learning and data science domains, including classification, natural language processing, and recommender systems. Among its several components, Figure \ref{fig:perfect} presents the categorical distribution of problems in DevAI. Each problem in Devai is not merely a question with a binary correct/incorrect outcome, but rather a structured task, decomposed into multiple requirements.

% \begin{table}[!h]
%     \centering
%     \caption{Category frequencies in the DevAI benchmark.}
%     \label{tab:devai_categories}
%     \begin{tabular}{l c}
%         \toprule
%         \textbf{Category} & \textbf{Frequency} \\
%         \midrule
%         Supervised Learning & 26 \\
%         Classification & 18 \\
%         Computer Vision & 14 \\
%         Natural Language Processing & 12 \\
%         Generative Models & 9 \\
%         Financial Analysis & 6 \\
%         Medical Analysis & 5 \\
%         Regression & 4 \\
%         Reinforcement Learning & 4 \\
%         Time Series Forecasting & 3 \\
%         Audio Processing & 3 \\
%         Unsupervised Learning & 3 \\
%         Recommender Systems & 2 \\
%         Other & 2 \\
%         \bottomrule
%     \end{tabular}
% \end{table}

As illustrated in Figure \ref{fig:task_51}, every problem is accompanied by: 1) The main question statement, which describes the problem to be solved. 2) A set of requirements, representing the individual steps, constraints, or conditions that a correct solution must satisfy. 3) A dependency graph, capturing the logical dependencies between requirements. Certain requirements can only be evaluated if prerequisite requirements have already been satisfied. 

\begin{figure}
    \centering
    \begin{tcolorbox}[
        colframe=blue,
        colback=white,
        coltitle=blue,
        title=\textbf{\textcolor{blue}{Example task: Tweets sentiment analysis on Sentiment140 dataset from HF}},
        fonttitle=\bfseries,
        boxrule=0.5mm,
        width=\columnwidth,
        halign=center,
        colbacktitle=blue!10,
        halign title=flush center,
        boxsep=2pt,
        left=2pt,
        right=2pt,
        top=2pt,
        bottom=2pt
    ]

    \begin{tcolorbox}[
        colframe=blue, colback=blue!3,
        sharp corners,
        boxrule=0.4mm,
        boxsep=2pt,
        left=2pt, right=2pt, top=2pt, bottom=2pt
    ] 
        \centering
        \textbf{\textcolor{blue}{Query}}\\
        \scriptsize
        \justifying
        \emph{Build a sentiment analysis system using the Sentiment140 dataset from Hugging Face. Load and clean the data (remove stop words, punctuation, special characters) in \texttt{src/data\_loader.py}. Use Word2Vec or GloVe for vectorization in the same file. Train an SVM classifier in \texttt{src/model.py} and save the accuracy in \texttt{results/metrics/accuracy\_score.txt}.}
    \end{tcolorbox}

    \textbf{\textcolor{blue}{Requirements}} 
    \scriptsize
    \begin{itemize}[left=0pt, itemsep=4pt, label={\textcolor{blue}{$\blacksquare$}}]
        \item \textbf{\textcolor{blue}{R0}} 
        \emph{Sentiment140 is loaded in \texttt{src/data\_loader.py}.}\\
        \textbf{Dependencies} \dependencyarrow \{\}

        \item \textbf{\textcolor{blue}{R1}} 
        \emph{Dataset is cleaned (stop words, punctuation, special characters) in \texttt{src/data\_loader.py}.}\\
        \textbf{Dependencies} \dependencyarrow \{\textbf{\textcolor{blue}{R0}}\}

        \item \textbf{\textcolor{blue}{R2}} 
        \emph{Word2Vec or GloVe embeddings applied in \texttt{src/data\_loader.py}.}\\
        \textbf{Dependencies} \dependencyarrow \{\textbf{\textcolor{blue}{R0, R1}}\}

        \item \textbf{\textcolor{blue}{R3}} 
        \emph{SVM model trained in \texttt{src/model.py}.}\\
        \textbf{Dependencies} \dependencyarrow \{\textbf{\textcolor{blue}{R0, R1, R2}}\}

        \item \textbf{\textcolor{blue}{R4}} 
        \emph{Accuracy written to \texttt{results/metrics/accuracy\_score.txt}.}\\
        \textbf{Dependencies} \dependencyarrow \{\textbf{\textcolor{blue}{R1, R2, R3}}\}
    \end{itemize}

    \end{tcolorbox}
    \caption{A task example in DevAI.}
    \label{fig:task_51}
\end{figure}

For the granular evaluation process, which assesses the quality of a solution produced by a candidate model, we employ OpenAI's \texttt{gpt-4o-mini}. For each requirement, the model provides a binary judgment (satisfied or unsatisfied), along with a natural language explanation justifying its decision.

Since DevAI does not provide official Ground-Truth solutions for its tasks, we constructed reference solutions. To ensure their reliability, we verified that each solution satisfied all predefined requirements using our granular evaluation framework prior to inclusion in experiments. As shown in Figure~\ref{fig:satisfaction_rates}, most tasks achieve 100\% requirement satisfaction. In a few cases, lower satisfaction scores occur due to two main factors: (1) some tasks rely on external datasets that are no longer publicly available, and (2) the LLM judge occasionally misclassifies correct outputs as unsatisfied due to limitations in understanding. Notably, this issue persists even with more capable judge models. However, even in cases with lower percentages, the absolute number of unsatisfied requirements is often small, usually a single missed requirement in tasks with few total requirements.

Our interactive evaluation experiments employ \texttt{gpt-4.1-mini} and \texttt{gpt-4o-mini} in separate evaluation runs, with each model serving independently as evaluator. The evaluator provides iterative feedback by analyzing the ground truth solution, the set of predefined requirements, the solution produced by the evaluated model (interviewee), as well as any errors encountered during execution in a Python interpreter. 

% Since the interactive process spans multiple iterative trials, the interviewer model maintains a history containing all previously generated solutions, encountered errors, and provided feedback. As this iterative process progresses, the accumulated history may eventually exceed the context window of the model. The same limitation applies to the interviewee model, which must also maintain awareness of prior interactions.

% To address this, when the context window approaches its limit, we perform a summarization step using the \texttt{gpt-4o} model. This summarization condenses the interaction history into a compact representation that preserves the essential information required for further evaluation. The summarized history then replaces the full context, allowing the evaluation process to continue uninterrupted.

For the interactive interviewer evaluator model, we set the temperature parameter to 0.3 to encourage responses that balance determinism and creativity, while ensuring a degree of consistency across repeated evaluations. The interviewee model uses the same temperature setting (0.3) for comparable behavior in solution generation. We configure the maximum token limit to 2000 tokens for the interviewer, allowing it to handle detailed feedback within each evaluation step, while permitting 5000 tokens for the interviewee to accommodate longer solutions to complex problems.

To ensure consistent behavior during interactive evaluation, we design a set of role-specific prompts for both the evaluator and the interviewee model. The evaluator is guided by a system prompt that defines its objectives and communication style, as well as an assistant prompt that specifies evaluation criteria, feedback strategies, and hinting procedures. The interviewee model receives a system prompt outlining its role, expected behavior, and response format, along with a detailed user prompt that directs its problem-solving approach and ensures adherence to task requirements. All prompts are provided in Appendix A.
\section{Evaluation}

\begin{figure}
    \centering
    \includegraphics[width=\linewidth]{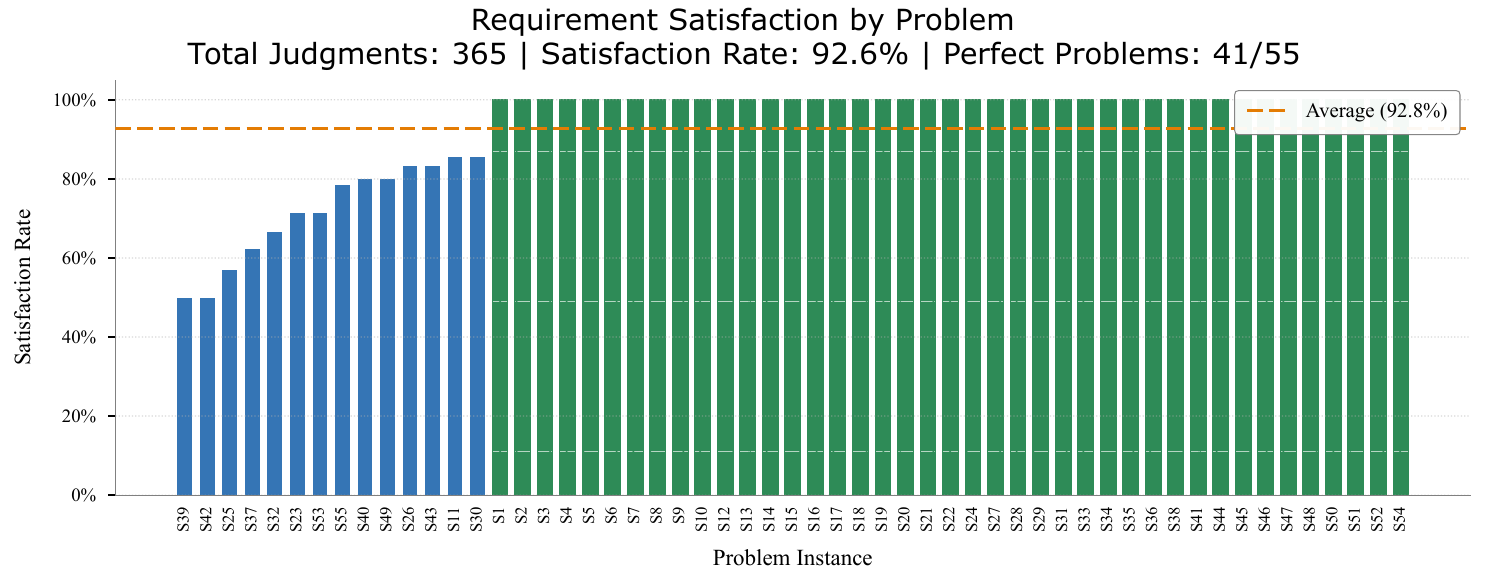}
    \caption{Requirement satisfaction rates of ground truths across all problems. Green bars indicate fully satisfied problems (100\%). The orange dashed line shows the average requirement satisfaction (92.6\%).}
    \label{fig:satisfaction_rates}
\end{figure}

We begin by rigorously evaluating the quality of our enhanced DevAI benchmark through multiple complementary analyses. First, we examine the requirement satisfaction rates of our curated Ground-Truth solutions. Figure \ref{fig:satisfaction_rates} reveals that 92.6\% of all requirements are satisfied on average across the benchmark, with a strong majority of problems achieving perfect 100\% compliance (shown in green). This high overall quality ensures that the interviewer model generates hints based on fundamentally sound reference implementations, establishing a solid foundation for reliable interactive evaluation.

Deeper category-level analysis in Figure \ref{fig:perfect} exposes important variations in solution quality across different software engineering domains. While well-structured tasks maintain near-perfect ground truth rates, more complex domains exhibit noticeable gaps. Specifically, generative models, computer vision, and NLP tasks demonstrate lower compliance rates. We attribute these differences to three key factors: (1) inherent ambiguity in problem specifications for creative tasks, (2) dependency on external data sources that may become unavailable, and (3) greater implementation variability in cutting-edge domains where best practices are still evolving.

\begin{figure}[t!]
    \centering
    \includegraphics[width=\linewidth]{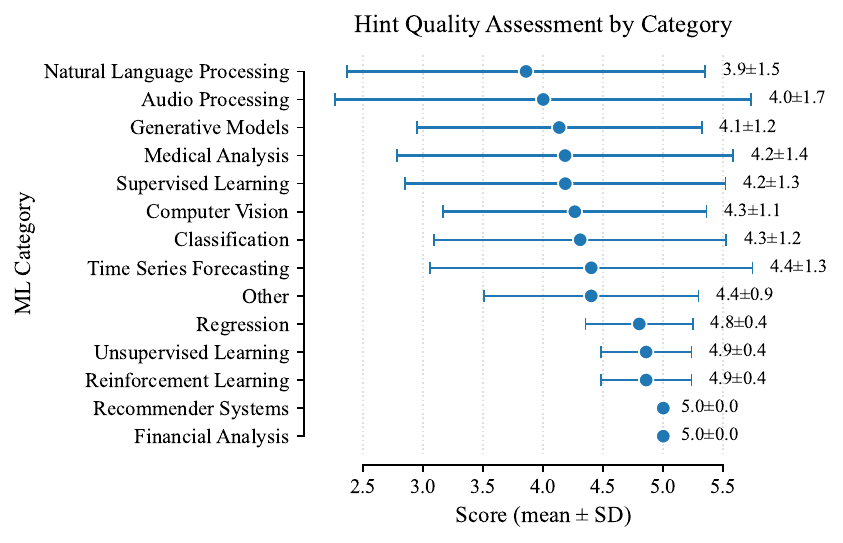}
    \caption{Hint quality scores across problem categories (mean = 4.32, $\sigma$ = 1.18). These hints were produced by GPT-4.1-mini.}
    \label{fig:hints_user_study_assessment}
\end{figure}

To assess how these benchmark characteristics translate to actual interactive evaluation quality, we conducted a comprehensive user study with 100 expert-annotated hints sampled from real evaluation sessions (20 hints $\times$ 5 interviewee models) using GPT-4.1-mini as the interviewer. The results in Figure \ref{fig:hints_user_study_assessment} demonstrate that overall hint quality remains strong ($\mu$=4.32/5, $\sigma$=1.18). We observe the predicted correlation between ground truth quality and hint effectiveness: where in categories with lower ground truth quality, we have slightly lower scores and hints showed greater variability. We replicated this study with GPT-4o-mini as the interviewer, with complete results and comparative analysis presented in Appendix C, revealing consistent patterns in hint effectiveness across both interviewer models.

To set baseline expectations, we begin by contextualizing model capabilities using OpenAI's published benchmark results \cite{gpt-4-1, o3-o4-mini, o3-mini}. The GPT-4.1-mini outperforms the GPT-4o-mini in traditional coding and instruction evaluations, scoring 24\% versus 9\% on SWE-bench, 35\% compared to 4\% on Aider’s Polyglot benchmark \cite{aider-leaderboards}, and 84\% against 78\% on IFEval \cite{zhou2023instructionfollowingevaluationlargelanguage}. Notably, the o3-mini surpasses both variants in standalone coding assessments with a 42.9\% score on SWE-bench, while the o4-mini exhibits comprehensive superiority across all major benchmarks.

\begin{figure}[t!]
    \centering
    \includegraphics[width=\linewidth]{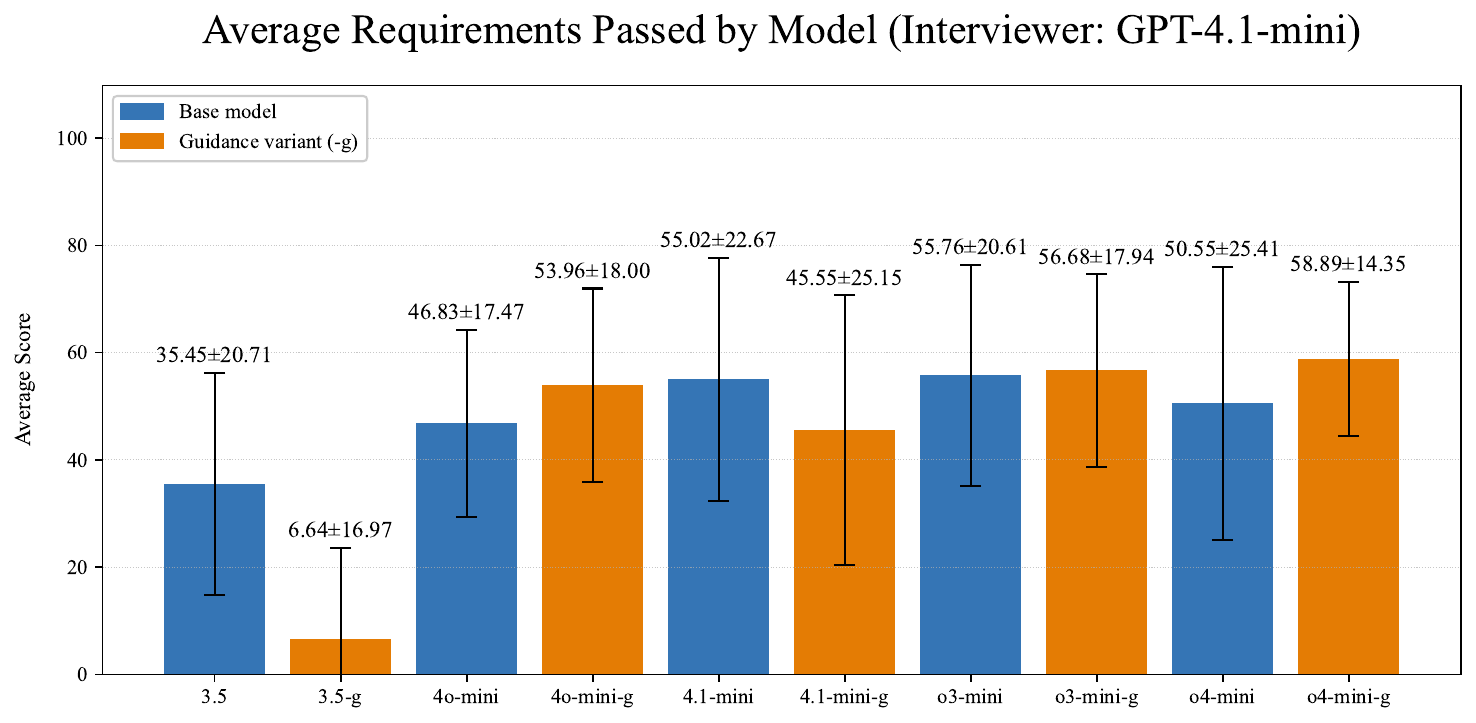}
    \caption{Average requirements passed by model variant using GPT-4.1.mini as interviewer. Blue bars represent base models, orange bars show guided variants (-g).}
    \label{fig:averages}
    
    \vspace{0.5cm} % Add some vertical spacing between figures
    
    \centering
    \includegraphics[width=\linewidth]{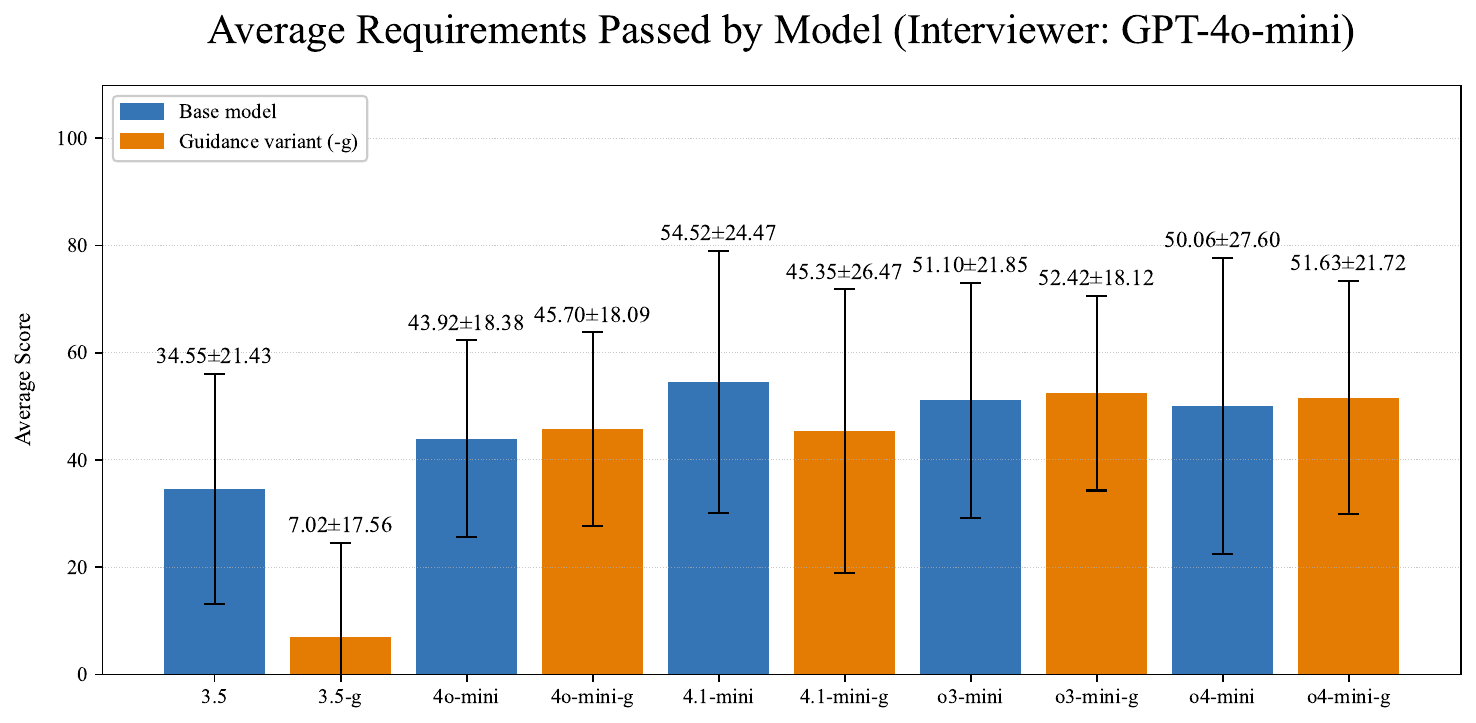}
    \caption{Average requirements passed by model variant using GPT-4o-mini as interviewer. Blue bars represent base models, orange bars show guided variants (-g). Interviewer: GPT-4o-mini}
    \label{fig:averages_}
\end{figure}

\begin{figure}[t!]
    \centering
    \includegraphics[width=\linewidth]{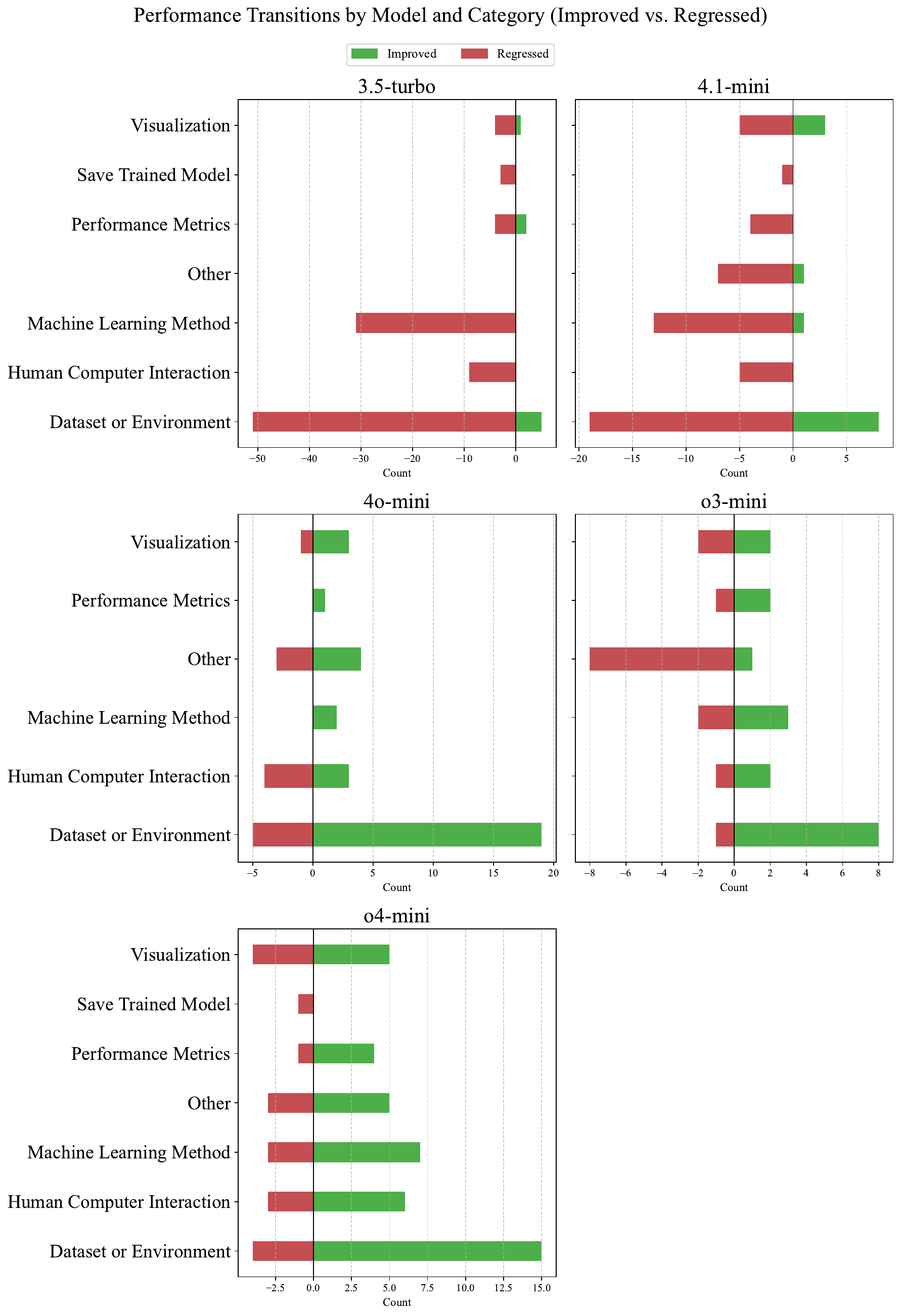}
    \caption{Breakdown of guidance impact across model variants per requirement category}
    \label{fig:ver_trans}
\end{figure}

These static benchmark results present an intriguing paradox when contrasted with our interactive evaluation findings. In the context of complex, multi-requirement software engineering tasks requiring iterative feedback and refinement, in Figure \ref{fig:averages}, where GPT-4.1-mini acts as the interviwer, we observe that (interviewee) GPT-4.1-mini's performance degrades relative to GPT-4o-mini, with guided variants only matching GPT-4o-mini's baseline unguided performance. Furthermore, while o4-mini initially demonstrates suboptimal performance on certain problem categories, its capacity for instruction following becomes evident through the guidance process, ultimately surpassing all other model variants in final performance. This divergence suggests that GPT-4.1-mini exhibits limitations in processing and incorporating multi-turn feedback during refinement cycles, while o4-mini's robust instruction-following capabilities enable it to overcome initial implementation challenges (particularly those related to dataset and environment configuration, as detailed in Figure \ref{fig:ver_trans}) and achieve superior final results.

Figure \ref{fig:averages_} shows that most models achieve only marginal gains when using hints from GPT-4o-mini, suggesting these hints provide limited value. Notably, GPT-4.1-mini exhibits a similar pattern – its performance deteriorates when relying on such hints. This decline stems from two compounding issues: the hints' inherent weaknesses and the model's inability to effectively utilize iterative refinements. Ultimately, this combination produces worse results than when the model generates solutions independently.

The observed behavior of GPT-3.5-turbo aligns with expectations given its architectural limitations. As a smaller model with constrained context window size, it struggles to: (1) retain and apply past refinements across iterations, (2) consistently produce complete solutions when task complexity exceeds its capacity, and (3) reliably follow instructions to regenerate full solutions – a weakness also documented in Aider's Polyglot benchmark. These limitations manifest consistently regardless of the interviewer model (GPT-4o-mini or GPT-4.1-mini), confirming fundamental capability constraints rather than interviewer-specific effects.

The transition plot (Figure \ref{fig:ver_trans}) reveals distinct patterns in how models respond to hint interventions across task categories. All models exhibit their most pronounced performance improvements in the Dataset or Environment category, with consistent positive counts observed for every model variant. This trend likely stems from the inherent challenges posed by benchmark tasks requiring manipulation of recent or niche datasets, where models frequently lack sufficient pretraining exposure or precise location information. The availability of ground truth references in hints appears particularly effective for resolving such environment-specific ambiguities. Beyond this commonality, models demonstrate divergent response profiles to hinting. These differential responses suggest that hint efficacy depends both on the task domain and the specific model's capability profile, with no universal improvement pattern emerging across the evaluated categories.

\section{Conclusion}

Our work establishes a new paradigm for evaluating LLMs in software engineering tasks through three fundamental contributions. First, we demonstrate that dependency-aware interactive evaluation reveals capabilities and limitations obscured by static benchmarks - where models like GPT-4.1-mini show unexpected performance degradation when processing iterative feedback, while o4-mini leverages its superior instruction follow-up capacity to overcome initial implementation challenges. Second, we enhance with Ground-Truths and validate the DevAI benchmark, and we expose how error recovery in later stages (e.g., model training) often compensates for early failures (e.g., data loading) when guided by targeted hints. Third, our findings challenge the prevailing assumption that benchmark performance directly translates to interactive settings, as evidenced by the weak correlation between static scores and guided improvement rates across model variants.

Several limitations warrant consideration. First, requirement extraction inherits ambiguities from natural language specifications, potentially introducing noise in task decomposition. Second, variability in automated feedback generation introduces a subtle but important fairness consideration in model comparisons. While our interviewer model generates hints consistently for all evaluated models, the effectiveness of these hints may vary based on each model’s architectural strengths. Most critically, the framework must carefully balance guidance intensity: overly specific hints risk revealing solutions, while vague suggestions may fail to trigger meaningful improvements.

\bibliography{aaai2026}

\newpage
\clearpage
\section*{Appendix}

\subsection*{A. Prompt Templates} \label{appendix:A}

This appendix provides the complete collection of prompt templates employed in our interactive evaluation framework, systematically documenting their roles and design principles. The prompts are categorized according to their function within the evaluation pipeline, distinguishing between system-level instructions that govern model behavior, user-initiated task specifications that define problem parameters, and assistant-generated responses that structure the interaction flow. Each prompt has been carefully crafted to maintain consistency across evaluation sessions while allowing for necessary flexibility in task-specific adaptations. 

\subsubsection{Initial solution construction}

The initial solution generation phase establishes the baseline implementation that subsequent interactive evaluation will refine. This process involves two critical prompt components: (1) a system prompt that configures the interviewee model's behavior as a coding assistant, specifying general constraints on response format and technical approach (Fig \ref{fig:unguided_system_interviewee}); and (2) a user instruction prompt that precisely structures the solution output to facilitate automated workspace generation (Fig \ref{fig:unguided_instruction}). Together, these prompts ensure solutions adhere to both functional requirements and our framework's parsing conventions while maintaining natural coding practices. 

\begin{figure}[h]
    \centering
    \begin{tcolorbox}[
        colframe=black,
        colback=red!30,
        coltitle=black,
        title=\textbf{\textcolor{black}{System Prompt for Interviewee (Unguided)}},
        fonttitle=\bfseries,
        boxrule=0.4mm,
        width=\columnwidth,
        colbacktitle=blue!10,
        halign title=flush center,
        boxsep=2pt,
        left=2pt,
        right=2pt,
        top=2pt,
        bottom=2pt
    ]
    \scriptsize
    \justifying
    \emph{You are a skilled coding interviewee. Provide complete, functional code solutions to programming problems. Include clear comments to highlight key parts of the solution. Do not include explanations—only the code.}
    \end{tcolorbox}
    \caption{System prompt provided to the interviewee for initial solution generation}
    \label{fig:unguided_system_interviewee}
\end{figure}

\begin{figure}
    \centering
    \begin{tcolorbox}[
        colframe=black,
        colback=red!30,
        coltitle=black,
        title=\textbf{\textcolor{black}{Interviewee instruction prompt (Unguided)}},
        fonttitle=\bfseries,
        boxrule=0.4mm,
        width=\columnwidth,
        colbacktitle=blue!10,
        halign title=flush center,
        boxsep=2pt,
        left=2pt,
        right=2pt,
        top=2pt,
    ]
    \scriptsize
    \justifying
Follow these steps strictly for the upcoming problem:

1. Understand the problem: Grasp the requirements, constraints, and expected output. Resolve any ambiguities using standard coding assumptions.

2. Plan the solution: Organize your approach for modularity, efficiency, and readability.

3. Write the code: Output each time the COMPLETE solution with properly structured files. Separate each file with a header in a code block, e.g.:

\texttt{```python}\\
\texttt{\# filename.py}\\
\texttt{(code)}\\
\texttt{```}

or

\texttt{```python}\\
\texttt{\# directory/filename.py}\\
\texttt{(code)}\\
\texttt{```}

4. For each runtime-generated file (e.g., logs, results, models, metrics), include a plaintext block (outside code blocks) specifically for this file, with its path and a note stating it will be created at runtime:

\texttt{```plaintext}\\
\texttt{\# filename1.extension}\\
\texttt{\# This file will be created at runtime}\\
\texttt{```}

\texttt{```plaintext}\\
\texttt{\# directory/filename2.extension}\\
\texttt{\# This file will be created at runtime}\\
\texttt{```}

5. Ensure completeness: Include all necessary imports, function definitions, and components.

6. No explanations: Provide only the code, with inline comments clearly referencing each corresponding requirement.

7. Modularize: Implement each requirement in its own function.

8. Follow best practices: Write clean, well-documented, maintainable code that handles edge cases.

9. Assume a fully prepared environment: Do not include dependency installations or directory creation commands; focus solely on functionality. However, don't assume datasets are readily available; you may need to install them if needed.

10. Also, include a file named execute\_workspace.sh that runs all components in sequence, ensuring the code executes correctly and produces all required outputs. Use the following format:

\texttt{```bash}\\
\texttt{\# execute\_workspace.sh}\\
\texttt{(code)}\\
\texttt{```}

Output only the structured code with proper file separation and no extra commentary.
    \end{tcolorbox}
    \caption{User prompt provided to the interviewee for the guidelines}
    \label{fig:unguided_instruction}
\end{figure}

\subsubsection{Interactive Guided Evaluation}

In this subsection, we describe the structure of prompts used for the Interactive Guided Evaluation protocol. The interaction begins with a system prompt for the interviewer, setting its role and responsibilities (Fig \ref{fig:guided_system_interviewer}). This is followed by an assistant prompt instructing the interviewer to critically evaluate a specific problem instance using a reference solution and a predefined set of evaluation guidelines (Fig \ref{fig:interviewer_guidelines}). Within this context, the interviewer is also made aware of its evaluative responsibilities and communicates this understanding explicitly, including a message directed at the interviewee stating the problem to be solved (Fig \ref{fig:akn}). Next, we define the system prompt for the interviewee, which provides general behavioral guidelines, especially regarding how to interpret and respond to feedback (Fig \ref{fig:system_int}). The user prompt to the interviewer then presents the initial solution proposed by the interviewee. Following this, a user prompt to the interviewee provides detailed instructions on how to respond thoughtfully and constructively during the evaluation (same as Fig \ref{fig:unguided_instruction}). The user prompt to the interviewee is then repeated to formally present the problem along with any relevant instructions (Fig \ref{fig:present}), and the assistant prompt of the interviewee follows with the initial solution. Finally, a user prompt to the interviewer requests the final assessment report based on the interaction (Fig \ref{fig:report_prompt}).

\begin{figure}
    \centering
    \begin{tcolorbox}[
        colframe=black,
        colback=red!30,
        coltitle=black,
        title=\textbf{\textcolor{black}{Interviewee system prompt}},
        fonttitle=\bfseries,
        boxrule=0.4mm,
        width=\columnwidth,
        colbacktitle=blue!10,
        halign title=flush center,
        boxsep=2pt,
        left=2pt,
        right=2pt,
        top=2pt,
        bottom=2pt
    ]
    \scriptsize
    \justifying
    You are a highly skilled coding interviewee tasked with solving programming problems efficiently. Your role is to provide complete, structured, and fully functional solutions without explanations, or reasoning—only the code itself. If given feedback, modify the solution accordingly in order to fix potential errors and always give the complete solution. Have commenting in code where requirements are implemented. Maintain the specified format at all times.
    \end{tcolorbox}
    \caption{System prompt for the interviewee}
    \label{fig:system_int}
\end{figure}

\begin{figure}
    \centering
    \begin{tcolorbox}[
        colframe=black,
        colback=red!30,
        coltitle=black,
        title=\textbf{\textcolor{black}{Interviewer System Prompt}},
        fonttitle=\bfseries,
        boxrule=0.4mm,
        width=\columnwidth,
        colbacktitle=blue!10,
        halign title=flush center,
        boxsep=2pt,
        left=2pt,
        right=2pt,
        top=2pt,
        bottom=2pt
    ]
    \scriptsize
    \justifying
    You are a technical interviewer specialized in evaluating coding and problem-solving skills of a candidate model. Your goal is to provide precise, minimal, and structured feedback, strictly addressing the requirements of the problem presented.

        Always follow these evaluation rules:

        1. Requirement-Oriented: Explicitly reference the provided requirements and criteria.  
        
        2. Dependency-Aware: Consider requirement dependencies; if a prerequisite requirement is unmet, prioritize hints addressing that first.
        
        3. Minimal and Incremental: Provide the minimal hint necessary for the candidate to identify their mistake.
        
        4. Objective and Specific: Clearly point out exactly one concrete issue per hint. Avoid vague or subjective feedback.
        
        5. Iterative Improvement: Assume multiple iterations. Guide incrementally without prematurely solving the entire task for the candidate.

        Your hints should be minimal, concise, and may include:
        - Conceptual pointers (e.g., "Verify the dimensions of your array.")
        - Specific references to requirements
        - Clarifying questions prompting the candidate to think critically.
    \end{tcolorbox}
    \caption{System prompt provided to the interviewer for the interactive evaluation}
    \label{fig:guided_system_interviewer}
\end{figure}

\begin{figure}
    \centering
    \begin{tcolorbox}[
        colframe=black,
        colback=red!30,
        coltitle=black,
        title=\textbf{\textcolor{black}{Interviewer assistant prompt guideline}},
        fonttitle=\bfseries,
        boxrule=0.4mm,
        width=\columnwidth,
        colbacktitle=blue!10,
        halign title=flush center,
        boxsep=2pt,
        left=2pt,
        right=2pt,
        top=2pt,
        bottom=2pt
    ]
    \scriptsize
    \justifying
    Problem for that I will be evaluating:

[START OF PROBLEM]
{problem}
[END OF PROBLEM]

Reference Solution for Guidance:

[START OF REFERENCE SOLUTION]
{reference\_solution}
[END OF REFERENCE SOLUTION]

Evaluation Guidelines:

1. I will assess the given solution strictly based on the problem requirements without revealing my reasoning. I will:

   - Verify correctness, logic, and adherence to constraints.
   
   - Ensure all stated requirements are met (IMPORTANT).
   
   - Check that each requirement is implemented with an explicit inline comment linking it to the corresponding requirement.
   
   - Not introduce or evaluate any unstated requirements.

2. If the solution meets all requirements and is executed without errors, I will immediately respond with: "INTERVIEW IS OVER."

3. If the solution is incomplete or partially correct, I will provide one concise paragraph of minimal hints based on the reference solution:

   - I will focus solely on improvements based on the stated requirements.
   
   - I will avoid asking for execution details, test cases, outputs, or explanations.
   
   - I WILL NOT ASK FOR EXECUTION OF THE SCRIPT.
   
   - I will request explicit inline comments that reference each specific requirement.
   
   - I will not offer compliments (e.g., “Good job” or “Well done”).
   
   - I will provide hints informed by the reference solution—unknown to the interviewee—to guide their improvements. If the solution remains uncorrected and the same error persists, I will progressively reveal more explicit hints based on the reference solution. If the mistake repeats, I will provide code snippets from the reference solution to steer them toward the correct approach.

4. I WILL NOT ASK THE INTERVIEWEE TO RUN THE CODE.

5. IMPORTANT: I WILL PROVIDE the snippet from the reference solution that downloads the dataset for the problem.

I will assume a fully prepared execution environment with all required packages installed. But I won't assume the datasets are readily available; the interviewee may need to install them.
    \end{tcolorbox}
    \caption{Assistant prompt for the interviewer for the guidelines}
    \label{fig:interviewer_guidelines}
\end{figure}

\begin{figure}
    \centering
    \begin{tcolorbox}[
        colframe=black,
        colback=red!30,
        coltitle=black,
        title=\textbf{\textcolor{black}{Interviewer acknowledgment prompt}},
        fonttitle=\bfseries,
        boxrule=0.4mm,
        width=\columnwidth,
        colbacktitle=blue!10,
        halign title=flush center,
        boxsep=2pt,
        left=2pt,
        right=2pt,
        top=2pt,
        bottom=2pt
    ]
    \scriptsize
    \justifying
    Understood. Now I will address the user. I will now act as technical interviewer and guide through the evaluation process.

        The problem we will examine is as follows:

        [START OF PROBLEM]
        {problem}
        [END OF PROBLEM]
    \end{tcolorbox}
    \caption{Assistant prompt for the interviewer acknowledging the evaluation process and stating the problem}
    \label{fig:akn}
\end{figure}

\begin{figure}
    \centering
    \begin{tcolorbox}[
        colframe=black,
        colback=red!30,
        coltitle=black,
        title=\textbf{\textcolor{black}{Interviewee problem user prompt}},
        fonttitle=\bfseries,
        boxrule=0.4mm,
        width=\columnwidth,
        colbacktitle=blue!10,
        halign title=flush center,
        boxsep=2pt,
        left=2pt,
        right=2pt,
        top=2pt,
        bottom=2pt
    ]
    \scriptsize
    \justifying
[START OF PROBLEM]
        {query}
        [END OF PROBLEM]

        From now on, I will provide feedback on your solution. After receiving feedback, please adjust your code accordingly in order to correct it, focusing on correctness, efficiency, and clarity. Provide complete, structured, and fully functional solutions without explanations, or reasoning—only the code itself. At all times maintain the specified format. Always give the FULL SOLUTION, not just the modifications.
    \end{tcolorbox}
    \caption{User prompt to interviewee for presenting the problem}
    \label{fig:present}
\end{figure}

\begin{figure}
    \centering
    \begin{tcolorbox}[
        colframe=black,
        colback=red!30,
        coltitle=black,
        title=\textbf{\textcolor{black}{Report prompt}},
        fonttitle=\bfseries,
        boxrule=0.4mm,
        width=\columnwidth,
        colbacktitle=blue!10,
        halign title=flush center,
        boxsep=2pt,
        left=2pt,
        right=2pt,
        top=2pt,
        bottom=2pt
    ]
    \scriptsize
    \justifying
Provide a structured assessment of my performance, focusing only on areas where I was incorrect or required hints. Include a detailed breakdown of the hints given, explaining their impact on my reasoning and how they guided the solution. Do not provide feedback on correct aspects of my solution—keep the evaluation concise and strictly focused on areas that needed improvement
    \end{tcolorbox}
    \caption{User prompt to interviewer for report}
    \label{fig:report_prompt}
\end{figure}

\subsection{B. Hints and additional statistics}

This appendix provides comprehensive supporting data from our interactive evaluation experiments. First, we present characteristic examples of generated hints (Figures \ref{fig:hint1}, \ref{fig:hint2}, \ref{fig:hint3}, \ref{fig:hint4}, \ref{fig:hint5}. \ref{fig:hint6}, \ref{fig:hint7}, \ref{fig:hint8}, \ref{fig:hint9}, \ref{fig:hint10}, \ref{fig:hint11}, \ref{fig:hint12}, \ref{fig:hint13}, \ref{fig:hint14}, \ref{fig:hint15}) across different model configurations and problem categories, illustrating the variation in feedback quality and specificity. These examples demonstrate how hint formulation adapts to both the interviewee model's capabilities and the problem's technical requirements. The section concludes with a complementary visualization of model performance.

\begin{figure}
    \centering
    \begin{tcolorbox}[
        colframe=black,
        colback=brown!30,
        coltitle=black,
        title=\textbf{\textcolor{black}{Interviewee: o4-mini, Problem: S10}},
        fonttitle=\bfseries,
        boxrule=0.4mm,
        width=\columnwidth,
        colbacktitle=blue!10,
        halign title=flush center,
        boxsep=2pt,
        left=2pt,
        right=2pt,
        top=2pt,
        bottom=2pt
    ]
    \scriptsize
    \justifying
    \emph{Your code attempts to load the dataset using \texttt{fetch\_openml} with the name \texttt{heart-disease-uci}, but this dataset is not found on OpenML, causing a failure to load data. The requirement explicitly states to use the UCI Heart Disease dataset from the provided link, and the reference solution downloads and extracts it manually or via a Kaggle dataset. Please revise your \texttt{src/data\_loader.py} to download and load the dataset from the official source or a reliable mirror, ensuring the dataset is available at runtime as in the reference solution. Also, add explicit inline comments referencing the requirement for dataset loading.}
    \end{tcolorbox}
    \caption{Hint provided by the interviewer}
    \label{fig:hint1}
\end{figure}

\begin{figure}
    \centering
    \begin{tcolorbox}[
        colframe=black,
        colback=brown!30,
        coltitle=black,
        title=\textbf{\textcolor{black}{Interviewee: o4-mini, Problem: S10}},
        fonttitle=\bfseries,
        boxrule=0.4mm,
        width=\columnwidth,
        colbacktitle=blue!10,
        halign title=flush center,
        boxsep=2pt,
        left=2pt,
        right=2pt,
        top=2pt,
        bottom=2pt
    ]
    \scriptsize
    \justifying
    \emph{Your data preprocessing pipeline in \texttt{src/data\_loader.py} does not handle categorical string features before applying \texttt{SelectKBest} and \texttt{StandardScaler}, causing a failure when converting strings like \texttt{normal} to float. The reference solution converts all features to numeric and removes or encodes categorical columns before feature selection and scaling. Please ensure you preprocess or encode categorical features to numeric types before feature selection and scaling, and add inline comments referencing the requirement for feature selection and data standardization. This is a prerequisite before training the model and generating SHAP values}
    \end{tcolorbox}
    \caption{Hint provided by the interviewer}
    \label{fig:hint2}
\end{figure}

\begin{figure}
    \centering
    \begin{tcolorbox}[
        colframe=black,
        colback=brown!30,
        coltitle=black,
        title=\textbf{\textcolor{black}{Interviewee: o4-mini, Problem: S10}},
        fonttitle=\bfseries,
        boxrule=0.4mm,
        width=\columnwidth,
        colbacktitle=blue!10,
        halign title=flush center,
        boxsep=2pt,
        left=2pt,
        right=2pt,
        top=2pt,
        bottom=2pt
    ]
    \scriptsize
    \justifying
    \emph{Your solution correctly downloads the dataset from a reliable source, encodes categorical features, applies feature selection and standardization with clear inline comments referencing requirements. The model training script implements both XGBoost and Logistic Regression with performance metrics, saves the XGBoost model under \texttt{models/saved\_models/}, and generates SHAP explanations saved in the required paths. The ROC curve is saved under \texttt{results/figures/roc\_curve.png}, and the HTML report includes all required visualizations and performance comparisons, saved as \texttt{results/report.html}. Inline comments explicitly link code sections to requirements. This meets all stated requirements. }
    \end{tcolorbox}
    \caption{Hint provided by the interviewer}
    \label{fig:hint3}
\end{figure}

\begin{figure}
    \centering
    \begin{tcolorbox}[
        colframe=black,
        colback=brown!30,
        coltitle=black,
        title=\textbf{\textcolor{black}{Interviewee: o4-mini, Problem: S30}},
        fonttitle=\bfseries,
        boxrule=0.4mm,
        width=\columnwidth,
        colbacktitle=blue!10,
        halign title=flush center,
        boxsep=2pt,
        left=2pt,
        right=2pt,
        top=2pt,
        bottom=2pt
    ]
    \scriptsize
    \justifying
    \emph{Your solution must include a script named \texttt{execute\_workspace.sh} that runs the full pipeline as described: first data preprocessing and visualization, then training and evaluation, saving outputs to the specified paths. Please add this script with explicit inline comments referencing the requirement to run preprocessing, training, inference, and saving results and models accordingly.}
    \end{tcolorbox}
    \caption{Hint provided by the interviewer}
    \label{fig:hint4}
\end{figure}

\begin{figure}
    \centering
    \begin{tcolorbox}[
        colframe=black,
        colback=brown!30,
        coltitle=black,
        title=\textbf{\textcolor{black}{Interviewee: o4-mini, Problem: S38}},
        fonttitle=\bfseries,
        boxrule=0.4mm,
        width=\columnwidth,
        colbacktitle=blue!10,
        halign title=flush center,
        boxsep=2pt,
        left=2pt,
        right=2pt,
        top=2pt,
        bottom=2pt
    ]
    \scriptsize
    \justifying
    \emph{Your current \texttt{download\_and\_extract\_enron} function attempts to download the dataset from a URL but encounters SSL certificate verification failure. To comply with the requirement to load the Enron-Spam dataset efficiently and reliably, consider using the \texttt{kagglehub} library as in the reference solution, which downloads the dataset from Kaggle with built-in handling and avoids manual URL and SSL issues. This also ensures the dataset is loaded in \texttt{src/data\_loader.py} as required. Additionally, add explicit inline comments referencing the requirement about dataset loading and preprocessing for clarity.}
    \end{tcolorbox}
    \caption{Hint provided by the interviewer}
    \label{fig:hint5}
\end{figure}

\begin{figure}
    \centering
    \begin{tcolorbox}[
        colframe=black,
        colback=brown!30,
        coltitle=black,
        title=\textbf{\textcolor{black}{Interviewee: o4-mini, Problem: S38}},
        fonttitle=\bfseries,
        boxrule=0.4mm,
        width=\columnwidth,
        colbacktitle=blue!10,
        halign title=flush center,
        boxsep=2pt,
        left=2pt,
        right=2pt,
        top=2pt,
        bottom=2pt
    ]
    \scriptsize
    \justifying
    \emph{The DataFrame column name for labels is incorrect. Instead of \texttt{"Category"}, use \textit{"Spam/Ham"} as in the reference solution. Update your \texttt{load\_data()} function to map labels from \texttt{df["Spam/Ham"]} to 1 for \texttt{"spam"} and 0 for \texttt{"ham"}. This fixes the KeyError and aligns with the requirement to load and preprocess the Enron-Spam dataset correctly. Also, add an explicit inline comment referencing this requirement}
    \end{tcolorbox}
    \caption{Hint provided by the interviewer}
    \label{fig:hint6}
\end{figure}

\begin{figure}
    \centering
    \begin{tcolorbox}[
        colframe=black,
        colback=brown!30,
        coltitle=black,
        title=\textbf{\textcolor{black}{Interviewee: o4-mini, Problem: S45}},
        fonttitle=\bfseries,
        boxrule=0.4mm,
        width=\columnwidth,
        colbacktitle=blue!10,
        halign title=flush center,
        boxsep=2pt,
        left=2pt,
        right=2pt,
        top=2pt,
        bottom=2pt
    ]
    \scriptsize
    \justifying
    \emph{Your current code does not include explicit inline comments referencing each specific requirement, which is necessary for clarity and grading. Also, the dataset target column is named \texttt{Churn} in the reference solution, but you use \texttt{target} in \texttt{preprocess\_data}; please verify the exact column name in the loaded dataset. Additionally, your cross-validation uses \texttt{cross\_val\_score} on the training set only, but the problem requires cross-validation during training with predictions used for the classification report on the entire dataset. Consider using \texttt{cross\_val\_predict} with stratified folds as in the reference. Finally, your data loading function lacks a comment explicitly linking it to the requirement for error handling during dataset loading. Please add these inline comments and verify column names to align with the problem requirements.}
    \end{tcolorbox}
    \caption{Hint provided by the interviewer}
    \label{fig:hint7}
\end{figure}

\begin{figure}
    \centering
    \begin{tcolorbox}[
        colframe=black,
        colback=brown!30,
        coltitle=black,
        title=\textbf{\textcolor{black}{Interviewee: o3-mini, Problem: S26}},
        fonttitle=\bfseries,
        boxrule=0.4mm,
        width=\columnwidth,
        colbacktitle=blue!10,
        halign title=flush center,
        boxsep=2pt,
        left=2pt,
        right=2pt,
        top=2pt,
        bottom=2pt
    ]
    \scriptsize
    \justifying
    \emph{Your solution currently does not explicitly load the Electronics subset of the Amazon Reviews 2023 dataset using the \texttt{datasets} library as required. Instead, it reads from a local CSV or uses a dummy dataset fallback. To meet the requirement, please implement data loading in \texttt{src/data\_loader.py} using the \texttt{load\_dataset} function from the \texttt{datasets} library with the \texttt{"McAuley-Lab/Amazon-Reviews-2023"} dataset and \texttt{"raw\_review\_Electronics"} configuration, as shown in the reference solution. Also, please add explicit inline comments referencing the requirement for data loading and preprocessing steps.}
    \end{tcolorbox}
    \caption{Hint provided by the interviewer}
    \label{fig:hint8}
\end{figure}

\begin{figure}
    \centering
    \begin{tcolorbox}[
        colframe=black,
        colback=brown!30,
        coltitle=black,
        title=\textbf{\textcolor{black}{Interviewee: o3-mini, Problem: S26}},
        fonttitle=\bfseries,
        boxrule=0.4mm,
        width=\columnwidth,
        colbacktitle=blue!10,
        halign title=flush center,
        boxsep=2pt,
        left=2pt,
        right=2pt,
        top=2pt,
        bottom=2pt
    ]
    \scriptsize
    \justifying
    \emph{Your current matrix factorization training uses SGD with multiple epochs, which can be computationally expensive and may cause timeout issues. The requirement suggests applying a latent factor model via SVD on a user-item rating matrix after preprocessing (noise removal and normalization). Consider replacing the SGD training with the SVD-based matrix factorization approach as in the reference solution (\texttt{scipy.sparse.linalg.svds} on a demeaned rating matrix). This method is more efficient and aligns with the requirement to use a latent factor model for user-item interactions. Also, please add explicit inline comments referencing each requirement in your code.}
    \end{tcolorbox}
    \caption{Hint provided by the interviewer}
    \label{fig:hint9}
\end{figure}

\begin{figure}
    \centering
    \begin{tcolorbox}[
        colframe=black,
        colback=brown!30,
        coltitle=black,
        title=\textbf{\textcolor{black}{Interviewee: 4.1-mini, Problem: S14}},
        fonttitle=\bfseries,
        boxrule=0.4mm,
        width=\columnwidth,
        colbacktitle=blue!10,
        halign title=flush center,
        boxsep=2pt,
        left=2pt,
        right=2pt,
        top=2pt,
        bottom=2pt
    ]
    \scriptsize
    \justifying
    \emph{Your DCGAN discriminator architecture causes a runtime error due to a convolution kernel size larger than the input feature map size during forward pass. Verify the dimensions after each Conv2d layer to ensure the kernel size fits the input; the reference solution uses a simpler discriminator architecture with downsampling steps that reduce 28x28 to 7x7 before flattening, avoiding this issue. Also, your execute\_workspace.sh script tries to save the PDF report to a non-existent nested directory \texttt{notebooks/results/}; ensure the output path directory exists or adjust the path to save directly under \texttt{results/} as in the reference solution.}
    \end{tcolorbox}
    \caption{Hint provided by the interviewer}
    \label{fig:hint10}
\end{figure}

\begin{figure}
    \centering
    \begin{tcolorbox}[
        colframe=black,
        colback=brown!30,
        coltitle=black,
        title=\textbf{\textcolor{black}{Interviewee: 4.1-mini, Problem: S23}},
        fonttitle=\bfseries,
        boxrule=0.4mm,
        width=\columnwidth,
        colbacktitle=blue!10,
        halign title=flush center,
        boxsep=2pt,
        left=2pt,
        right=2pt,
        top=2pt,
        bottom=2pt
    ]
    \scriptsize
    \justifying
    \emph{Your updated download\_image function now properly sets a User-Agent header and handles HTTP errors, which aligns well with the requirement to reliably download the content and style images. However, the problem statement also requires saving the style-transferred images in \texttt{results/figures/} (not just the final output but also intermediate results), and logging processing time to \texttt{results/processing\_time.txt}. Please add explicit inline comments referencing these requirements where you save images and log processing time, to clearly demonstrate compliance. Also, ensure that the intermediate results are saved exactly as \texttt{results/figures/intermediate\_results.png} as required, and the final stylized image is saved in the same directory. Adding these inline comments will improve clarity and traceability to the problem requirements.}
    \end{tcolorbox}
    \caption{Hint provided by the interviewer}
    \label{fig:hint11}
\end{figure}

\begin{figure}
    \centering
    \begin{tcolorbox}[
        colframe=black,
        colback=brown!30,
        coltitle=black,
        title=\textbf{\textcolor{black}{Interviewee: 4o-mini, Problem: S17}},
        fonttitle=\bfseries,
        boxrule=0.4mm,
        width=\columnwidth,
        colbacktitle=blue!10,
        halign title=flush center,
        boxsep=2pt,
        left=2pt,
        right=2pt,
        top=2pt,
        bottom=2pt
    ]
    \scriptsize
    \justifying
    \emph{Your dataset download function in \texttt{src/data\_loader.py} is a step in the right direction, but it lacks the specific logic to download and extract the TuSimple dataset from Kaggle as required. The problem explicitly requires downloading \texttt{"manideep1108/tusimple"} via Kaggle (using \texttt{kagglehub} or equivalent), checking for the presence of \texttt{train\_set/} and \texttt{test\_set/} folders, and handling ZIP extraction or directory copying accordingly. Also, your augmentation pipeline should include random cropping, rotation, and scaling (random resized crop) as per the requirement, rather than fixed resizing after cropping and rotation. Additionally, your model's final layer outputs 1 unit, but the problem expects 2 classes (binary classification with 2 outputs) to use \texttt{CrossEntropyLoss}. Please revise your dataset download logic to match the reference solution's approach, adjust your augmentation transforms to include \texttt{RandomResizedCrop} and \texttt{RandomRotation}, and modify your model's final layer to output 2 classes with appropriate loss. Also, add explicit inline comments referencing each requirement in your code}
    \end{tcolorbox}
    \caption{Hint provided by the interviewer}
    \label{fig:hint12}
\end{figure}

\begin{figure}
    \centering
    \begin{tcolorbox}[
        colframe=black,
        colback=brown!30,
        coltitle=black,
        title=\textbf{\textcolor{black}{Interviewee: 4o-mini, Problem: S26}},
        fonttitle=\bfseries,
        boxrule=0.4mm,
        width=\columnwidth,
        colbacktitle=blue!10,
        halign title=flush center,
        boxsep=2pt,
        left=2pt,
        right=2pt,
        top=2pt,
        bottom=2pt
    ]
    \scriptsize
    \justifying
    \emph{Your \texttt{execute\_workspace.sh} script tries to install packages from \texttt{requirements.txt}, but this file is missing, causing the installation to fail and the \texttt{facenet\_pytorch} module not to be found. Please ensure that \texttt{requirements.txt} exists in the workspace root with the necessary packages listed, including \texttt{facenet-pytorch}, and that your script installs from it before running the training script. This will resolve the missing module error and meet the requirement to load the FaceNet model without issues.}
    \end{tcolorbox}
    \caption{Hint provided by the interviewer}
    \label{fig:hint13}
\end{figure}

\begin{figure}
    \centering
    \begin{tcolorbox}[
        colframe=black,
        colback=brown!30,
        coltitle=black,
        title=\textbf{\textcolor{black}{Interviewee: 3.5-turbo, Problem: S48}},
        fonttitle=\bfseries,
        boxrule=0.4mm,
        width=\columnwidth,
        colbacktitle=blue!10,
        halign title=flush center,
        boxsep=2pt,
        left=2pt,
        right=2pt,
        top=2pt,
        bottom=2pt
    ]
    \scriptsize
    \justifying
    \emph{Your \texttt{load\_data} function in \texttt{src/data\_loader.py} does not load the Telco Customer Churn dataset from the Hugging Face \texttt{scikit-learn/churn-prediction} dataset as required; instead, it reads from a CSV file path. Also, you need to include error handling around dataset loading as specified. Please revise \texttt{load\_data} to use \texttt{load\_dataset("scikit-learn/churn-prediction")} from the \texttt{datasets} library with appropriate try-except error handling, and add an explicit inline comment referencing the dataset loading requirement. This is a prerequisite before addressing feature engineering and imbalance handling}
    \end{tcolorbox}
    \caption{Hint provided by the interviewer}
    \label{fig:hint14}
\end{figure}

\begin{figure}
    \centering
    \begin{tcolorbox}[
        colframe=black,
        colback=brown!30,
        coltitle=black,
        title=\textbf{\textcolor{black}{Interviewee: 3.5-turbo, Problem: S52}},
        fonttitle=\bfseries,
        boxrule=0.4mm,
        width=\columnwidth,
        colbacktitle=blue!10,
        halign title=flush center,
        boxsep=2pt,
        left=2pt,
        right=2pt,
        top=2pt,
        bottom=2pt
    ]
    \scriptsize
    \justifying
    \emph{Your \texttt{load\_sentiment140\_dataset} function attempts to load the dataset via `pd.read\_csv` from a Hugging Face URL, but the Sentiment140 dataset should be loaded using the Hugging Face \texttt{datasets} library as per the requirement. Also, your code does not map the sentiment labels from (0,4) to (0,1) for binary classification, which is required. Additionally, your vectorization function uses a local Word2Vec model file \texttt{word2vec.model} which is not provided or mentioned; the requirement is to use Word2Vec or GloVe embeddings loaded programmatically (e.g., via \texttt{gensim.downloader}). Finally, your code lacks explicit inline comments referencing each requirement, which are requested for clarity. Please revise your data loading to use the Hugging Face \texttt{load\_dataset("sentiment140")} method, map labels properly, use a pre-trained embedding model loaded in code, and add inline comments referencing each requirement}
    \end{tcolorbox}
    \caption{Hint provided by the interviewer}
    \label{fig:hint15}
\end{figure}

\begin{figure}
    \centering
    \includegraphics[width=\linewidth]{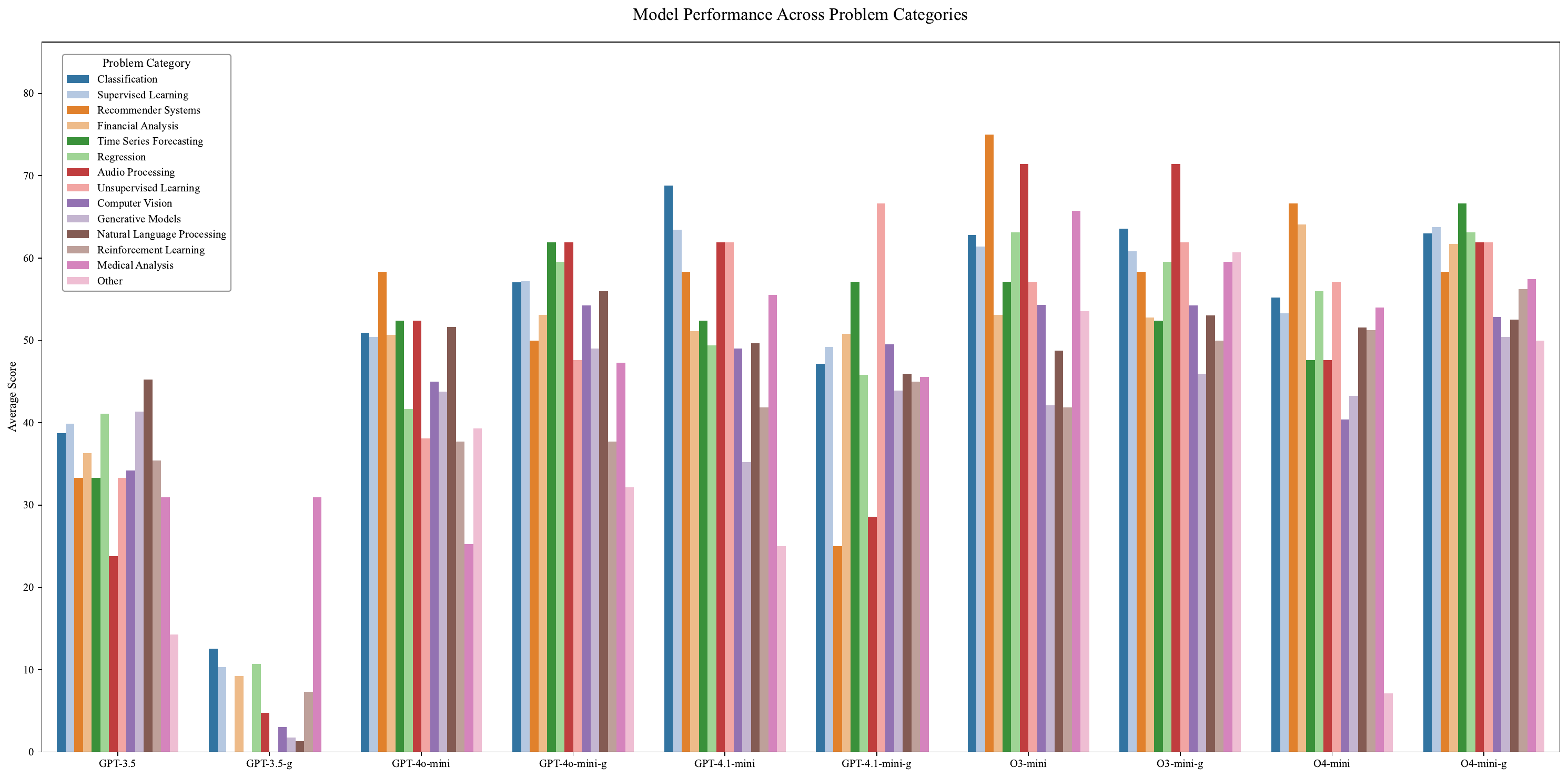}
    \caption{Performance across problem categories. Colors indicate different problem categories.}
    \label{fig:categories}
\end{figure}

Figure \ref{fig:categories} presents the performance distribution across problem categories for both guided and unguided model variants. The results reveal distinct capability profiles among models, with each variant exhibiting relative strengths in specific domains. Notably, the response to guidance varies significantly by both model architecture and problem category, demonstrating that hint effectiveness is context-dependent rather than uniform.

\begin{figure}
    \centering
    \includegraphics[width=0.8\linewidth]{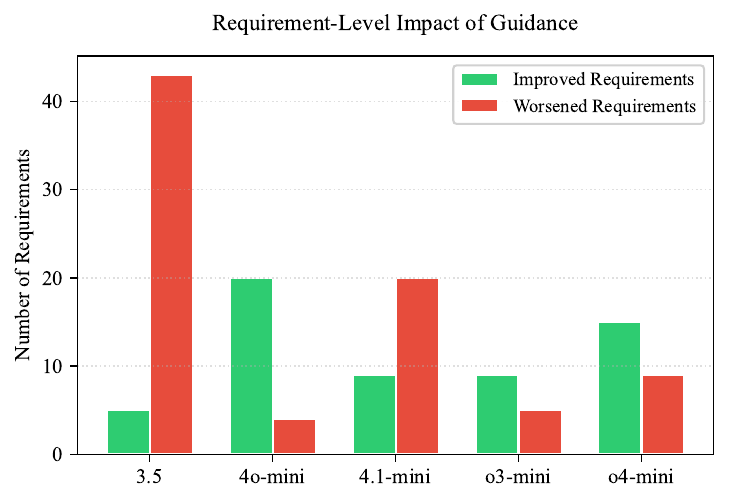}
    \caption{Breakdown of guidance impact across model variants. Bars show the number of problems where performance improved (green), worsened (red).}
    \label{fig:guidance_impact}
\end{figure}

Figure \ref{fig:guidance_impact} aggregates the per-category impact of guidance across all evaluated models, quantifying both improvements and regressions in requirement satisfaction. The visualization reveals model-specific response patterns to iterative feedback.

Figures \ref{fig:report1}, \ref{fig:report2}, and \ref{fig:report3} present examples of the evaluation reports generated during interactive guidance. These reports provide detailed analyses of model errors, the hints administered during interaction (including their success rates), task complexity assessments, and other key behavioral observations. Together, they offer a structured overview of the model’s strengths and weaknesses under guided evaluation.

\begin{figure}
    \centering
    \begin{tcolorbox}[
        colframe=black,
        colback=yellow!30,
        coltitle=black,
        title=\textbf{\textcolor{black}{Report - Interviewee: gpt-4.1-mini, Problem: S16}},
        fonttitle=\bfseries,
        boxrule=0.4mm,
        width=\columnwidth,
        colbacktitle=blue!10,
        halign title=flush center,
        boxsep=2pt,
        left=2pt,
        right=2pt,
        top=2pt,
        bottom=2pt
    ]
    \scriptsize
    \justifying
    1. **Initial Data Loading Method**: The candidate initially used `imdb.load\_data()` instead of the required `tfds.load()`. This indicated a lack of adherence to the requirement to use TensorFlow Datasets. The hint provided was to switch to using `tfds.load("imdb\_reviews", ...)`. This hint prompted the candidate to align their implementation with the specified requirements, showcasing their ability to adapt to feedback.
    
    2. **Embedding Saving Format**: The candidate initially saved embeddings using `pickle`, which was not specified in the requirements. The hint suggested saving the embeddings as a `.npy` file using `numpy`. This guidance helped the candidate correct the saving method, ensuring compliance with the requirements.
    
    3. **Model Architecture Visualization**: The initial implementation did not include saving the model architecture diagram. The hint emphasized the need to visualize and save the model architecture. This feedback led to the inclusion of the model architecture visualization, demonstrating the candidate's responsiveness to feedback.
    
    4. **Inline Comments**: The initial code lacked sufficient inline comments referencing specific requirements. The hint encouraged the addition of comments to clarify how each part of the code met the requirements. This improvement enhanced the overall organization and readability of the code.
    
    5. **Handling of Pre-trained Embeddings**: While the candidate used `gensim` to load embeddings, the hint focused on ensuring that the embeddings were correctly fetched and mapped. This guidance prompted the candidate to verify the loading process, which is critical for the model's performance.
    
    6. **Error Handling and Adaptability**: Throughout the iterations, the candidate demonstrated adaptability by addressing the hints provided and making necessary corrections. However, there was initially a lack of error handling in the code, particularly when loading embeddings. The hints did not explicitly address error handling, but the candidate's ability to implement changes based on feedback improved the robustness of the solution.
    
    Overall, the hints provided were targeted and specific, focusing on concrete issues that needed correction. Each hint had a direct impact on improving the solution, leading to a more compliant and organized implementation. The candidate's ability to adapt to feedback was evident in their incremental improvements, showcasing a logical approach to problem-solving.
    \end{tcolorbox}
    \caption{Report after the guided evaluation of gpt-4.1-mini by gpt-4o-mini on problem S16}
    \label{fig:report1}
\end{figure}

\begin{figure}
    \centering
    \begin{tcolorbox}[
        colframe=black,
        colback=yellow!30,
        coltitle=black,
        title=\textbf{\textcolor{black}{Report - Interviewee: gpt-4.1-mini, Problem: S37}},
        fonttitle=\bfseries,
        boxrule=0.4mm,
        width=\columnwidth,
        colbacktitle=blue!10,
        halign title=flush center,
        boxsep=2pt,
        left=2pt,
        right=2pt,
        top=2pt,
        bottom=2pt
    ]
    \scriptsize
    \justifying
    1. **Error Handling**: In `src/data\_loader.py`, the `preprocess\_image` function raises a `FileNotFoundError` if an image cannot be loaded. While this is a good practice, there is no handling for cases where the dataset path might be incorrect or if the directory structure is not as expected. A hint to improve this would be to add checks for the existence of the dataset directory and provide user-friendly error messages if the path is invalid. This would enhance robustness and user experience.

2. **Adaptability to Feedback**: The model's comments in the code are generally clear, but there was a lack of explicit references to how the output layer's activation function relates to the requirement for similarity scoring in `src/model.py`. A hint to improve this would be to add a comment explaining that the sigmoid activation function outputs a probability-like score, which is essential for determining if two images represent the same object. This would demonstrate a better understanding of the model's purpose.

3. **Complexity Awareness**: The model did not explicitly discuss time and space complexity in the implementation. For instance, in `generate\_pairs`, the function continuously generates pairs in an infinite loop without any mechanism to limit the number of iterations or to handle extremely large datasets efficiently. A hint to improve this would be to consider implementing a mechanism to limit the number of generated pairs or to allow for a configurable number of iterations, which would enhance both performance and usability.

4. **Code Organization**: While the code is structured well, there could be improvements in modularity. For example, the augmentation logic could be separated into its own class or module to enhance reusability and clarity. A hint to improve this would be to suggest creating a dedicated augmentation class that encapsulates all augmentation methods, making the codebase cleaner and more maintainable.

5. **Handling Ambiguity**: The model did not address how to handle potential ambiguities in the dataset, such as varying image sizes or formats. A hint to improve this would be to implement checks or preprocessing steps that standardize image sizes and formats before processing, ensuring consistency across the dataset.

Overall, the hints provided aimed to enhance error handling, adaptability to feedback, complexity awareness, code organization, and handling of ambiguities, which would collectively lead to a more robust and maintainable solution.
    \end{tcolorbox}
    \caption{Report after the guided evaluation of gpt-4.1-mini by gpt-4o-mini on problem S37}
    \label{fig:report2}
\end{figure}

\begin{figure}
    \centering
    \begin{tcolorbox}[
        colframe=black,
        colback=yellow!30,
        coltitle=black,
        title=\textbf{\textcolor{black}{Report - Interviewee: o3-mini, Problem: S8}},
        fonttitle=\bfseries,
        boxrule=0.4mm,
        width=\columnwidth,
        colbacktitle=blue!10,
        halign title=flush center,
        boxsep=2pt,
        left=2pt,
        right=2pt,
        top=2pt,
        bottom=2pt
    ]
    \scriptsize
    \justifying
    1. **Missing Function for Metadata Image Generation**: Initially, the solution did not include a function to generate an image with hidden text embedded in its metadata. The hint provided was to add this function, emphasizing the requirement to generate an image of 1080p resolution with hidden text in the metadata. This hint prompted the candidate to recognize the oversight and implement the necessary functionality, thereby aligning the solution with the problem requirements.

2. **Lack of Inline Comments for Metadata Function**: After the function for generating the metadata image was added, it lacked explicit inline comments referencing the specific requirements. The hint suggested adding comments to clarify how the function met the requirements. This feedback guided the candidate to improve the documentation of their code, enhancing readability and maintainability.

3. **Verification of Hidden Text**: There was an initial lack of emphasis on the requirement to manually verify that the hidden text was embedded in the generated images. The hint pointed out the need for comments that explicitly stated this verification step. This led to a more comprehensive understanding of the problem requirements and improved the clarity of the code regarding its functionality.

4. **Error Handling**: While the solution included some error handling (e.g., checking if the text length exceeded the number of available pixels), there was no indication of how the model would handle potential issues when reading or writing files (e.g., file permissions, nonexistent directories). A hint could have encouraged the candidate to consider adding error handling for file operations, which would improve the robustness of the solution.

5. **Complexity Awareness**: The solution did not address time and space complexity considerations, particularly in the context of embedding and extracting hidden text. A hint could have prompted the candidate to analyze the efficiency of their algorithms, especially since they involve manipulating pixel data in images. This would demonstrate a deeper understanding of performance implications in their solution.

6. **Code Organization**: The organization of the code could have been improved by separating concerns more clearly, such as grouping related functions or providing a clearer structure for the main execution flow. A hint suggesting better organization could have encouraged the candidate to refactor their code for improved clarity and maintainability.

Overall, the hints provided were instrumental in guiding the candidate to address these weaknesses, leading to a more complete and functional solution that adhered closely to the problem requirements. Each hint prompted the candidate to think critically about their implementation, ultimately enhancing their problem-solving skills and adaptability to feedback.
    \end{tcolorbox}
    \caption{Report after the guided evaluation of o3-mini by gpt-4o-mini on problem S8}
    \label{fig:report3}
\end{figure}

\subsubsection{C. User study}

\begin{figure}
    \centering
    \includegraphics[width=\linewidth]{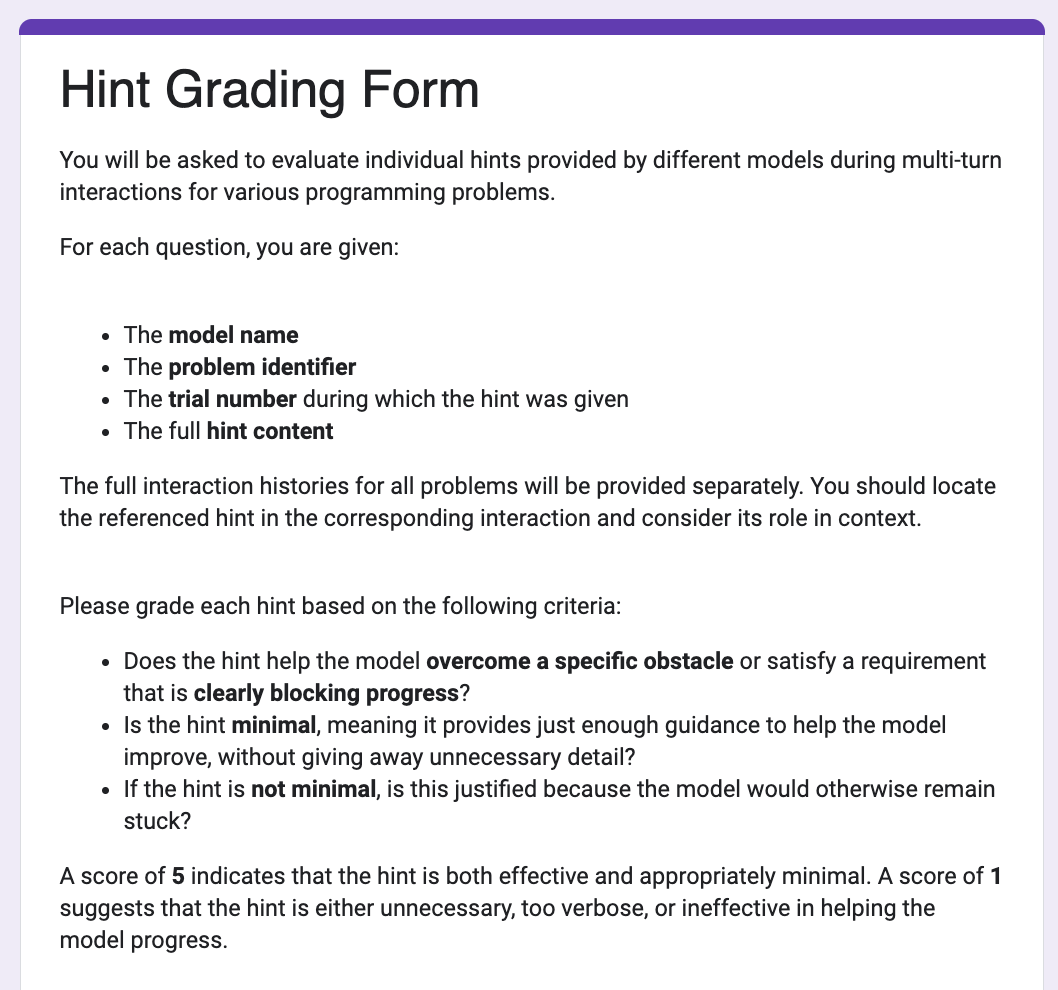}
    \caption{Description of user study}
    \label{fig:study_interface1}
\end{figure}

\begin{figure}
    \centering
    \includegraphics[width=\linewidth]{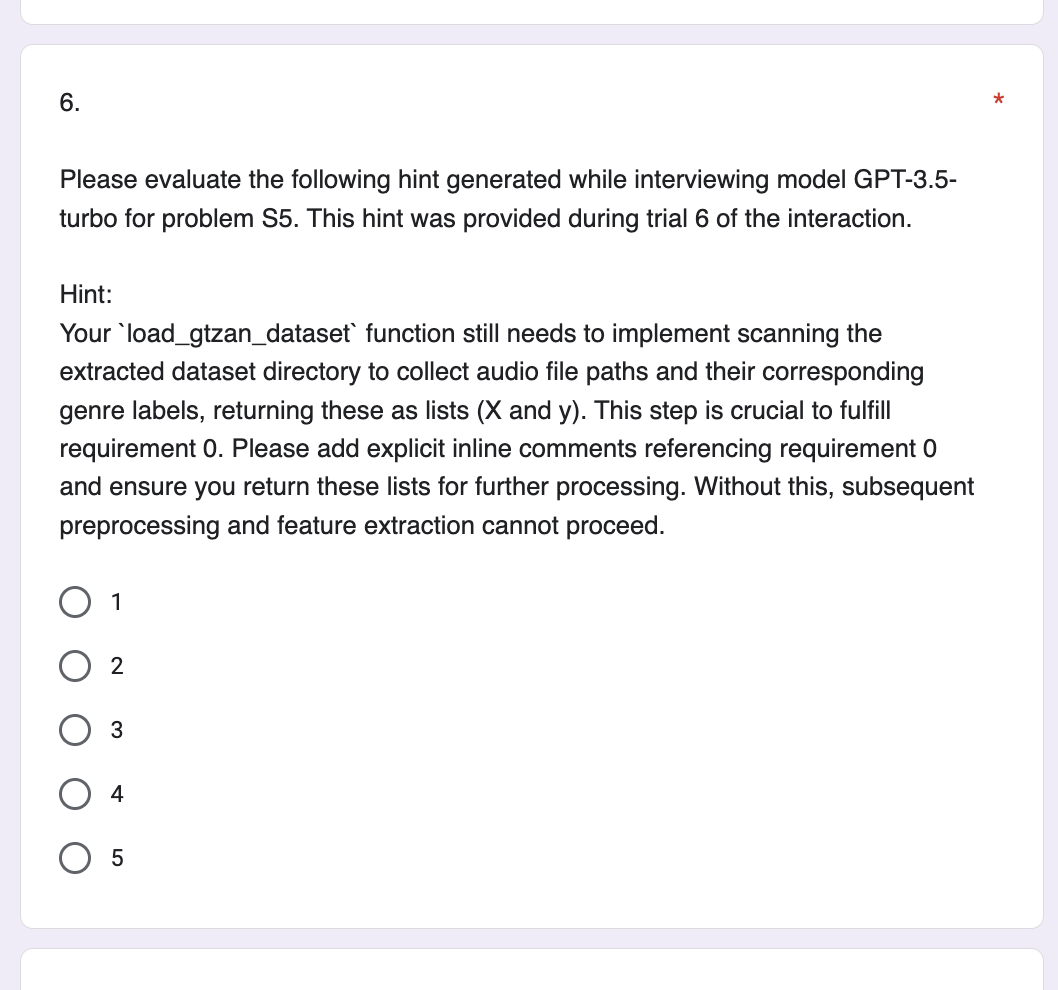}
    \caption{Hint example from user study}
    \label{fig:study_interface2}
\end{figure}

\begin{figure}
    \centering
    \includegraphics[width=\linewidth]{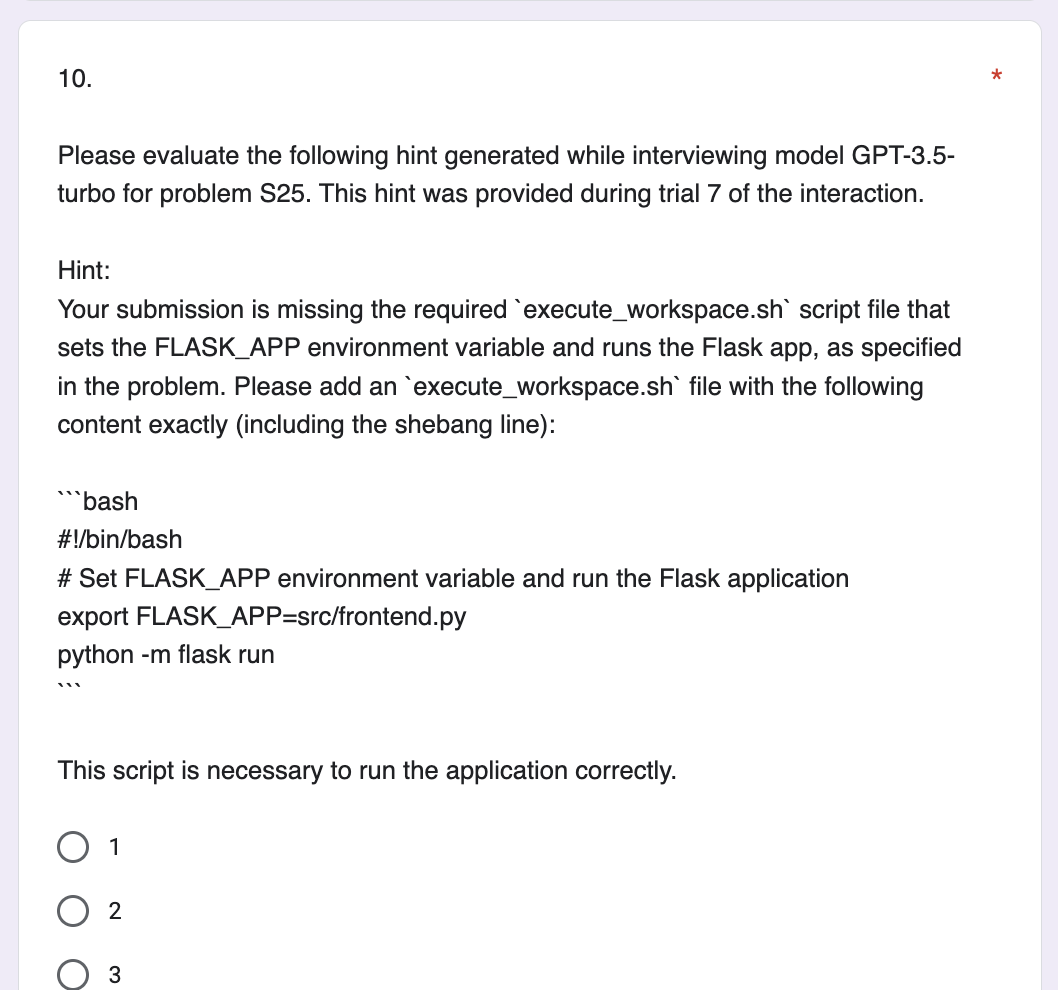}
    \caption{Hint example from user study}
    \label{fig:study_interface3}
\end{figure}

\begin{figure}
    \centering
    \includegraphics[width=\linewidth]{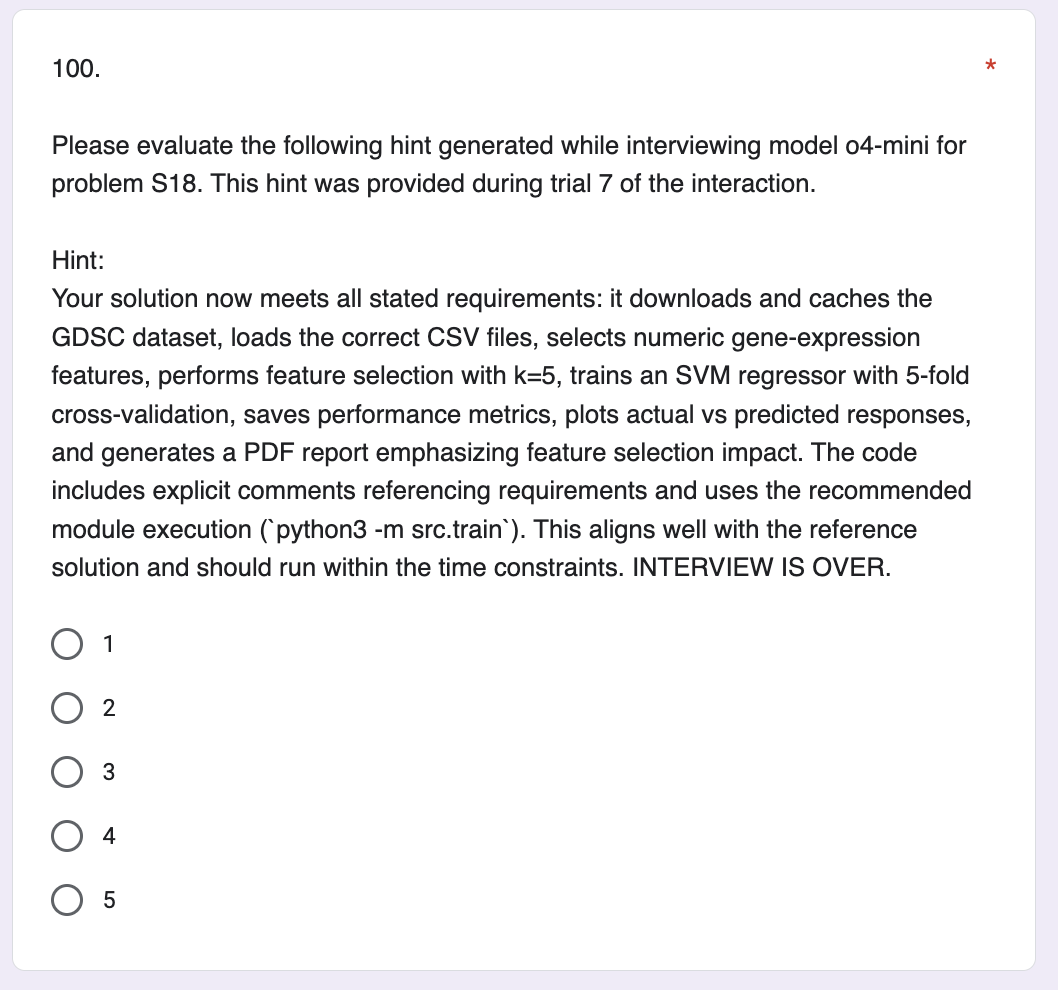}
    \caption{Hint example from user study}
    \label{fig:study_interface4}
\end{figure}

\begin{figure}
    \centering
    \includegraphics[width=\linewidth]{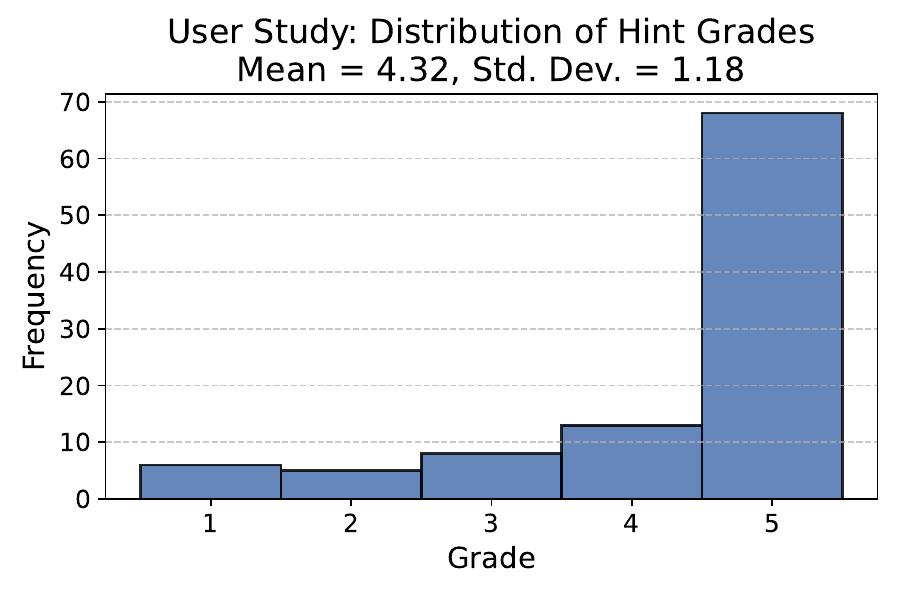}
    \caption{Distribution of Hint Grades (Interviewer GPT-4.1-mini)}
    \label{fig:hints_user_study}
\end{figure}

\begin{figure}
    \centering
    \includegraphics[width=\linewidth]{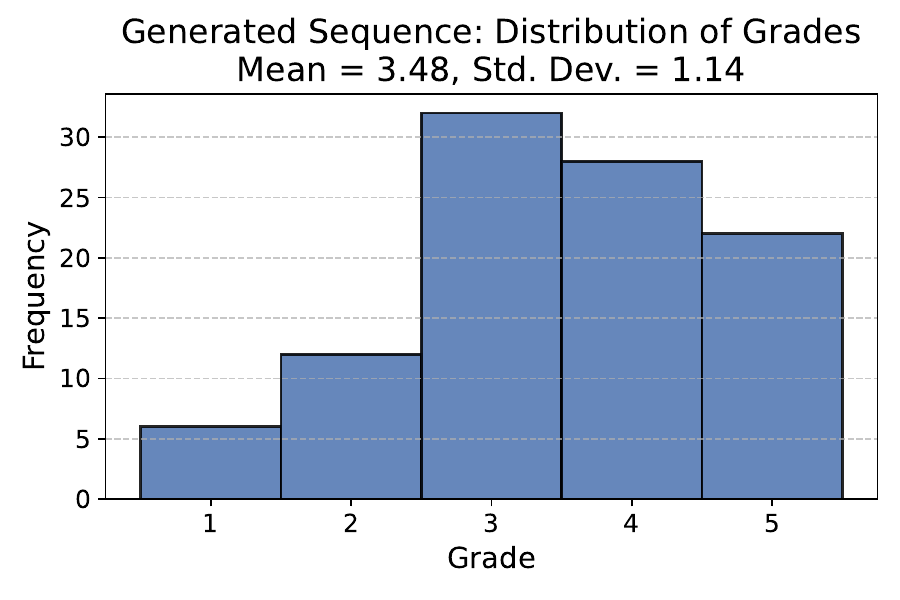}
    \caption{Distribution of Hint Grades (Interviewer GPT-4o-mini)}
    \label{fig:hints_user_study__}
\end{figure}

\begin{figure}
    \centering
    \includegraphics[width=\linewidth]{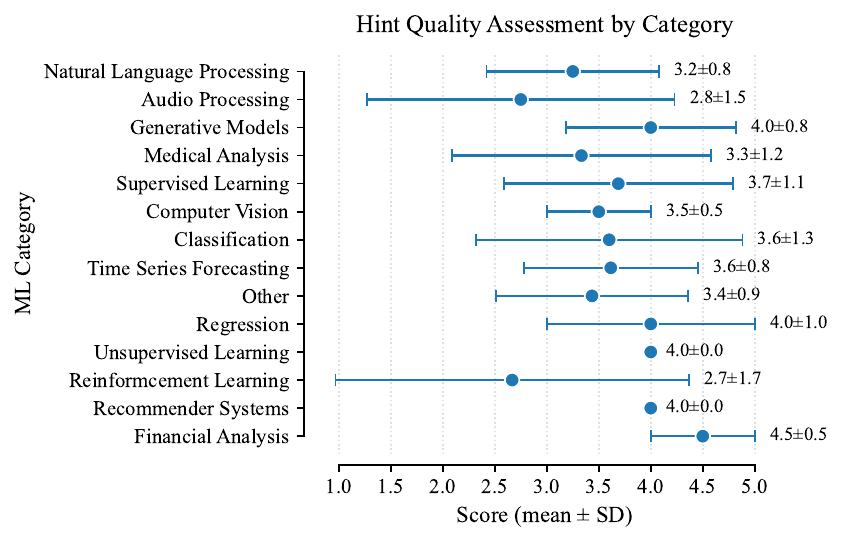}
    \caption{Hint quality scores across problem categories (mean = 3.48, $\sigma$ = 1.14). These hints were produced by GPT-4o-mini.}
    \label{fig:hints_user_cat_}
\end{figure}

Figures \ref{fig:study_interface1}, \ref{fig:study_interface2}, \ref{fig:study_interface3}, \ref{fig:study_interface4}  first illustrate the user study interface used for expert evaluations, demonstrating our annotation protocol for assessing hint quality. This section presents the resulting expert evaluations of hints generated by both interviewer models (GPT-4.1-mini and GPT-4o-mini). Figures \ref{fig:hints_user_study} and \ref{fig:hints_user_study__} present the expert-rated score distributions, while Figure \ref{fig:hints_user_cat_} breaks down scores by category. Both plots exhibit consistent patterns in hint quality across categories. However, the majority of scores cluster near the baseline (3/5), indicating that most hints were only moderately effective at advancing interactive evaluation. This limited efficacy likely explains the marginal improvements observed in guided model variants.

\end{document}